\documentclass[twocolumn]{svjour3}    
\usepackage{graphicx}
\usepackage{mathptmx}
\usepackage{amsmath}
\usepackage{nccmath}
\usepackage{cite}
\usepackage{latexsym}
\usepackage[numbers]{natbib}
\usepackage{graphicx}
\usepackage{cite}
\usepackage{algorithm}
\usepackage{amsmath,amssymb,amsfonts}
\usepackage{algorithmicx}
\usepackage{algpseudocode}
\usepackage{graphicx}
\usepackage{textcomp}
\usepackage{xcolor}
\usepackage{tikz}
\usepackage{pgfplots}
\usepackage{tablefootnote}
\definecolor{dgreen}{RGB}{0,130,0}
\begin{document}
\sloppy

\title{Subjective Annotation for a Frame Interpolation Benchmark using Artefact Amplification\thanks{Funded by the Deutsche Forschungsgemeinschaft (DFG, German Research Foundation) – Project-ID 251654672 – TRR 161 (Project A05 and B04).}}


\author{Hui Men$^{1}$ \and
Vlad Hosu$^{1}$ \and
Hanhe Lin$^{1}$ \and Andr{\'e}s Bruhn$^{2}$ \and Dietmar Saupe $^{1}$}


\institute{
$^1$ Department of Computer and Information Science, University of Konstanz, Germany. 
\email {\{hui.3.men, vlad.hosu, hanhe.lin, dietmar.saupe\}@uni-konstanz.de}
\\
$^2$
Institute for Visualization and Interactive Systems, University of Stuttgart, Germany. 
\email{bruhn@vis.uni-stuttgart.de }\\
}

\maketitle              

\begin{abstract}

Current benchmarks for optical flow algorithms evaluate the estimation either directly by comparing the predicted flow fields with the ground truth or indirectly by using the predicted flow fields for frame interpolation and then comparing the interpolated frames with the actual frames. In the latter case, objective quality measures such as the mean squared error are typically employed. However, it is well known that for image quality assessment, the actual quality experienced by the user cannot be fully deduced from such simple measures. Hence, we conducted a subjective quality assessment crowdscouring study for the interpolated frames provided by one of the optical flow benchmarks, the Middlebury benchmark. It contains interpolated frames from 155 methods applied to each of 8 contents. For this purpose, we collected forced-choice paired comparisons between interpolated images and corresponding ground truth. To increase the sensitivity of observers when judging minute difference in paired comparisons we introduced a new method to the field of full-reference quality assessment, called artefact amplification. From the crowdsourcing data (3720 comparisons of 20 votes each) we reconstructed absolute quality scale values according to Thurstone's model. As a result, we obtained a re-ranking of the 155 participating algorithms w.r.t.\ the visual quality of the interpolated frames. This re-ranking not only shows the necessity of visual quality assessment as another evaluation metric for optical flow and frame interpolation benchmarks, the results also provide the ground truth for designing novel image quality assessment (IQA) methods dedicated to perceptual quality of interpolated images. As a first step, we proposed such a new full-reference method, called WAE-IQA. By weighing the local differences between an interpolated image and its ground truth WAE-IQA performed slightly better than the currently best FR-IQA approach from the literature.

\end{abstract}


\section{Introduction}

As one of the basic video processing techniques, frame interpolation, namely computing interpolated in-between images in image sequences, is a necessary step in numerous applications such as temporal up-sampling for generating slow-motion videos \citep{jiang2018super}, nonlinear video re-timing in special effects movie editing \citep{mccann2006retiming}, and frame rate conversion between broadcast standards \citep{meyer2015phase}. One of the main concepts in frame interpolation is motion compensation. In this context, required frames are obtained by interpolating the image content along the path of motion. Thereby, the apparent motion in terms of the so-called optical flow can be derived in various ways. Typical approaches for this task include block matching techniques \citep{ha2004motion}, frequency-based approaches \citep{meyer2015phase}, variational methods \citep{raket2012motion} or convolutional neural networks \citep{memcnet,jiang2018super}.

Since the quality of the interpolated frames heavily depends on the underlying optical flow algorithm, the evaluation of the results is a critical issue. However, currently, there is only one optical flow benchmark that offers the assessment of interpolated frames: the Middlebury benchmark \citep{baker2011database}. Regarding the quality of the motion estimation, it considers angular and endpoint errors between the estimated flow and the ground truth flow. More importantly, it also offers a direct evaluation of the corresponding motion-compensated interpolation results between frame\linebreak pairs, which is based on the root mean squared error (RMSE) and the gradient normalized RMSE between the interpolated image and the ground truth image.

\begin{figure}[!t]
\centering{\includegraphics[width=0.48\textwidth]{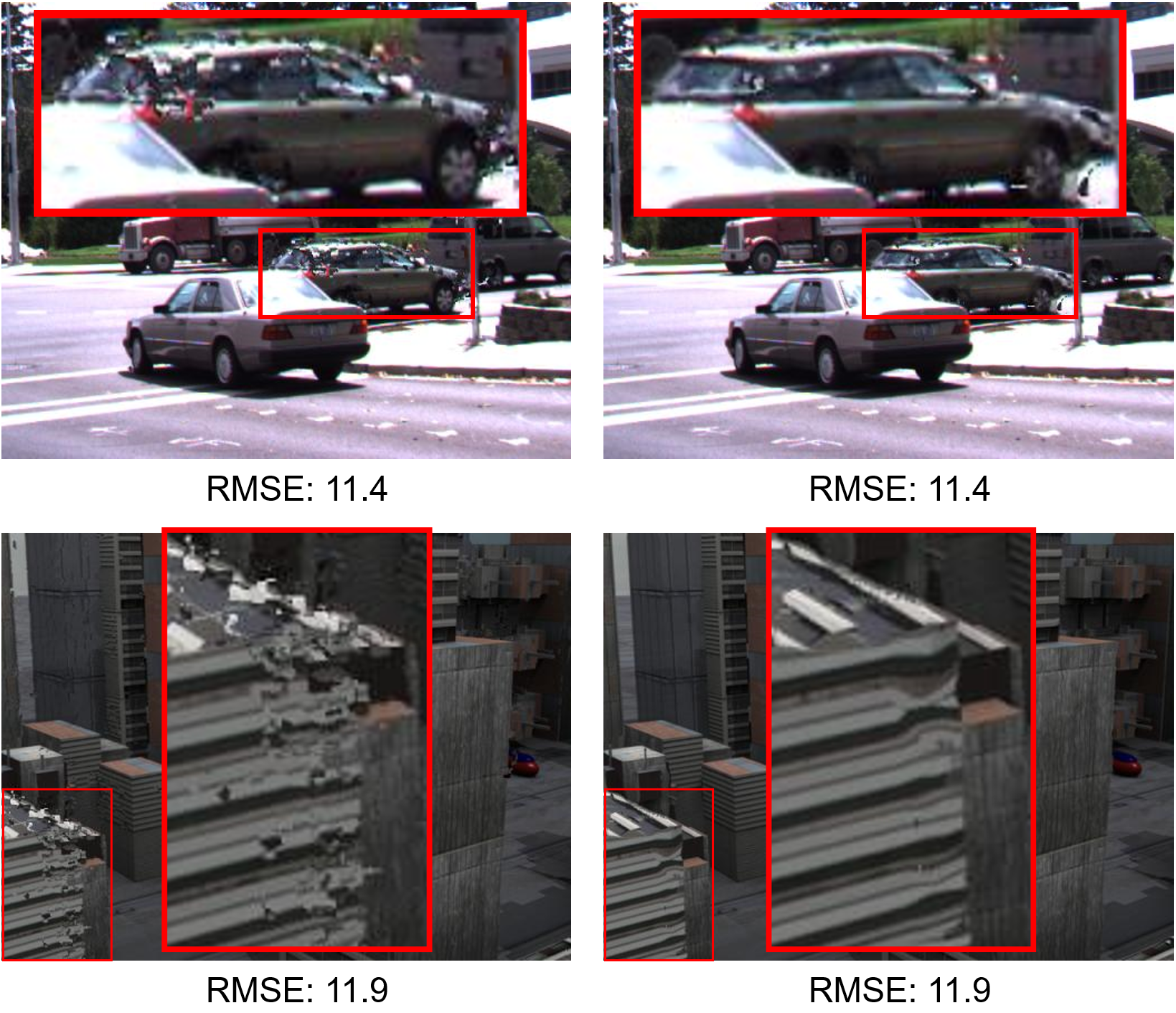}}
\vspace{-6pt}
\caption{Two interpolated frames, each with two different methods. RMSE values in each pair are equal, but the visual quality differs in each pair, in particular in the zoomed regions.
}
\label{MSE}
\vspace{-15pt}
\end{figure}

In general, the direct evaluation of the frame interpolation results is useful, since the accuracy of the motion estimation is not always directly related to the quality of the motion-compensated frame interpolation. For example, motion errors in low-textured regions are less noticeable in the interpolation result than motion errors in highly-textured regions. However, specifically designed error measures such as the gradient normalized RMSE even revert this relation and penalize interpolation errors in high-textured regions less severely which adapts the error measure to the shortcomings of motion-based frame interpolation techniques instead of trying to assess the frame interpolation quality adequately. Moreover, it is well known that even the standard mean square error can be misleading and may not reliably reflect image quality as perceived by the human visual system (HVS) \citep{wang2004image}. This fact also becomes obvious from the Middlebury web-page, where some of the interpolated images have the same RMSE but exhibit obvious differences in image quality (see Fig.~\ref{MSE}). Evidently, there is a clear need to improve the assessment of motion-compensated interpolation results. Therefore, we propose to change the quality assessment in such a way that the evaluation of the results takes perceived visual quality assessment into consideration.

\begin{figure}[t]
\centering{\includegraphics[width=0.48\textwidth]{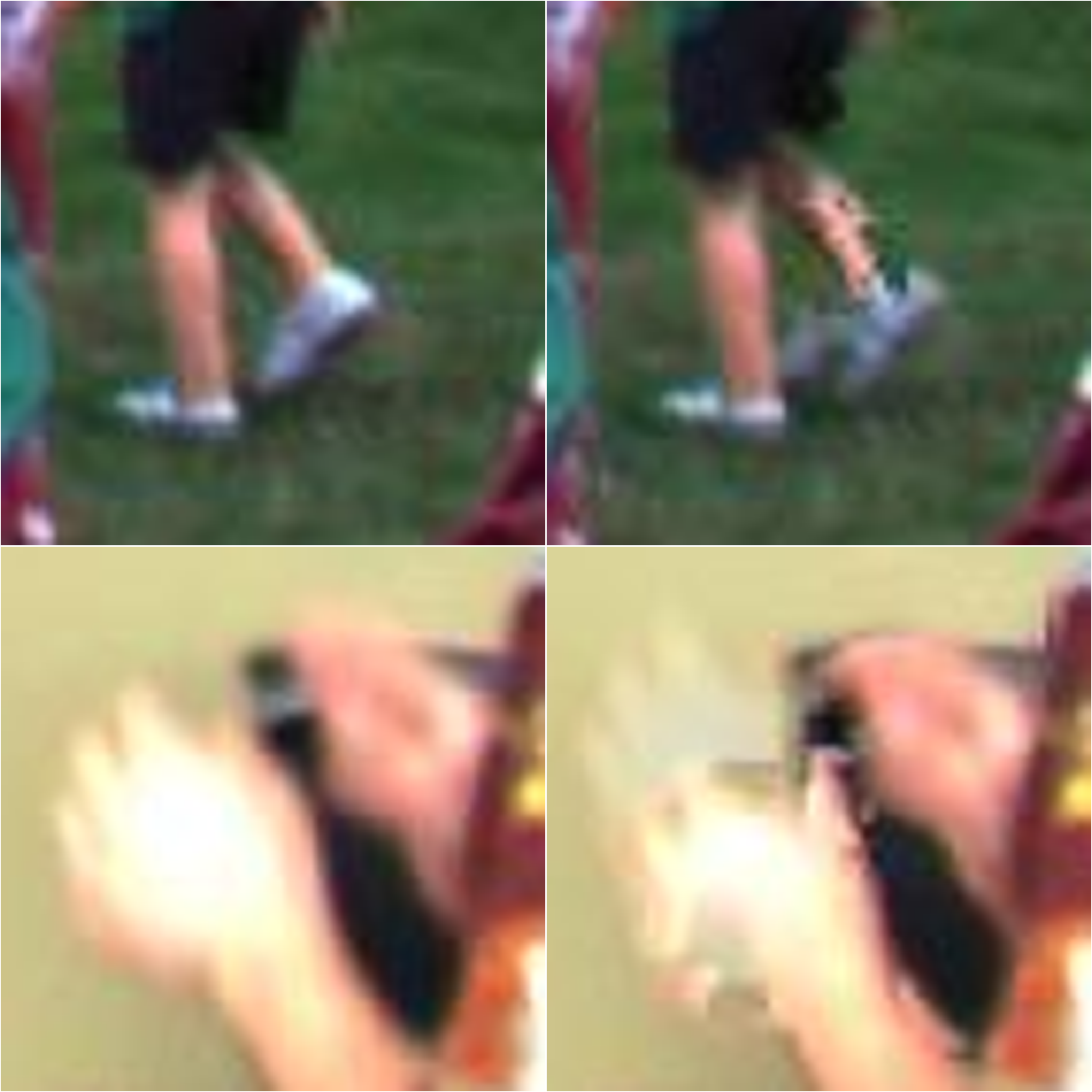}}
\caption{Specific distortions in interpolated images. 
Left column:
detail of the ground truth images \emph{Backyard} and \emph{Basketball}.
Right column: corresponding detail of the interpolated counterparts (with
distortions).}
\label{speci}
\end{figure}

Regarding visual quality assessment methods, we take full-reference image quality assessment (FR-IQA) into consideration, since ground truth in-between images are available in the Middlebury benchmark. There are several FR-IQA methods that consider the HVS, 
which were designed to estimate image quality degradation due to common artefacts, namely the ones caused by processing such as data compression or by losses in data transmission. However, the artefacts induced by optical flow algorithms lead to interpolated images with different specific distortions (see Fig.\ \ref{speci}).

In this article we show that nine of the most popular objective FR-IQA methods have rather low correlations with the evaluations made by human observers, regardless of whether the methods are based on the HVS or just on pixel-wise errors such as RMSE; see Table~\ref{tb:crd-iqa} (in Section \ref{sec:wae}). The VSI \citep{zhang2014vsi} method is one of the best FR-IQA methods. When trained and tested on the LIVE database it yields a Spearman rank-order correlation coefficient (SROCC) of 0.952 w.r.t.\ ground truth consisting of mean opinion scores (MOS) from a controlled lab study \citep{sheikh2006statistical}. However, even VSI only obtained an SROCC of 0.5397 w.r.t.\ MOS reconstructed from paired comparisons when applied to the interpolated images by optical flow algorithms.
This demonstrates that current FR-IQA methods are not able to cope with the specific distortion types that arise in interpolated frames produced by optical flow algorithms. Therefore, a new FR-IQA method specifically designed for such images is needed. However, before the research in such FR-IQA methods can proceed, ground truth data, namely subjective quality scores of such images need to be collected, a first set of which is provided by our study. 

Regarding the subjective quality evaluation, lab-studies are well established due to their reliability. In particular, the experimental environment and the evaluation process can be controlled. However, it is time-consuming and costly, which severely limits the number of images to be assessed. In contrast, crowdsourcing studies can be less expensive. Moreover, the reliability of crowdsourcing has been proven to be acceptable, if the results are properly post-processed by outlier removal \cite{ribeiro2011crowdsourcing}.

\begin{figure}[t]
\centering{\includegraphics[width=0.49\textwidth]{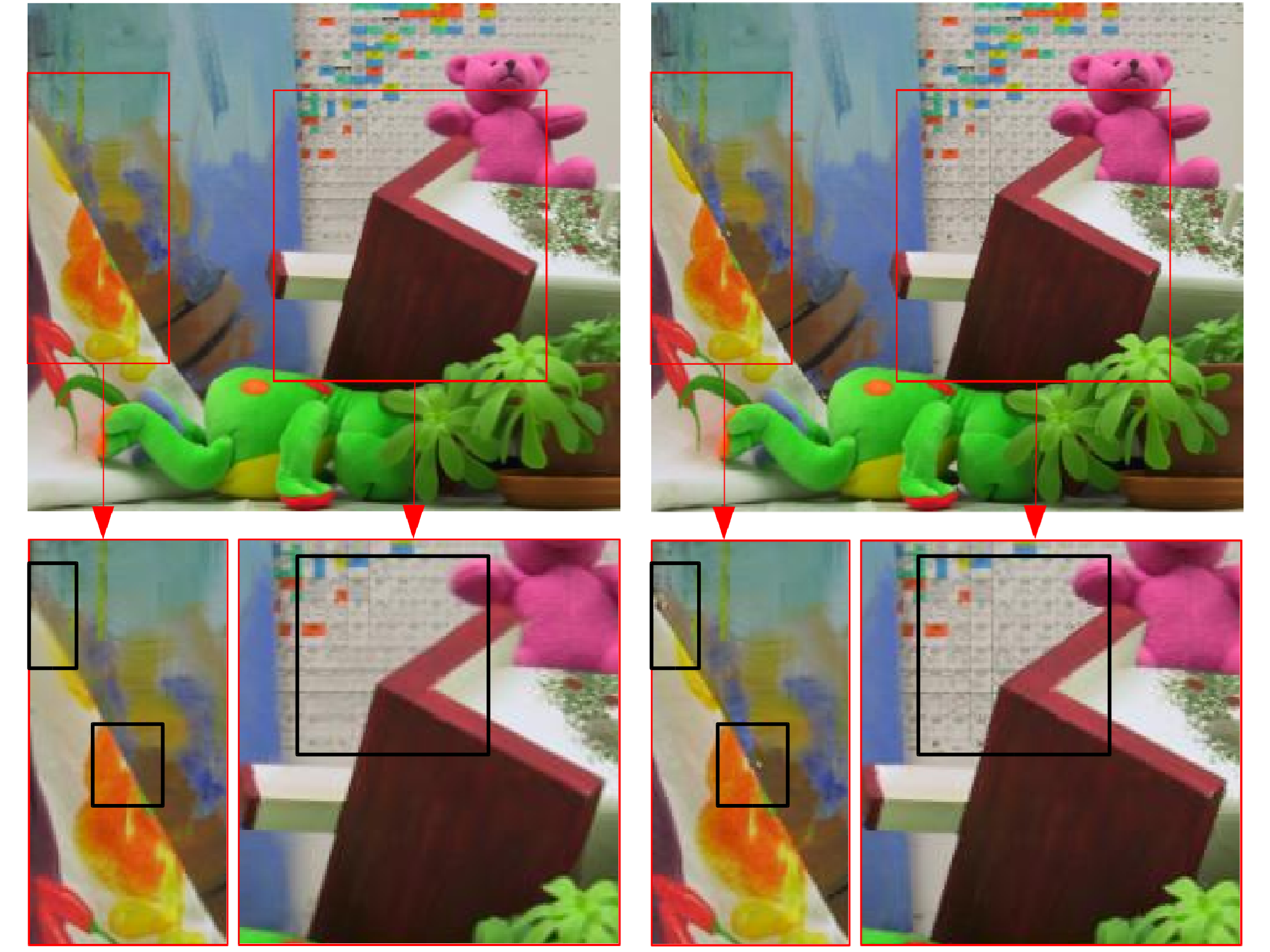}}
\caption{A pair of images that obtained the same average score in StudyMB~1.0. However, quality differences exist, especially in the zoomed-in parts (see the portions in black rectangles).}
\label{samevote_1}
\end{figure}

In our preliminary work, \citep{huiqomex2019}, with the help of crowdsourcing, we have implemented paired comparisons to collect subjective quality scores of the interpolated images in the Middlebury benchmark, denoted as \emph{StudyMB 1.0}. However, the limitation of StudyMB 1.0 is that the quality differences between some of the images are hardly visible. Even though in the instructions we had highlighted the main degraded parts according to our visual observation, still some of the image pairs could not be well judged by the subjects. For instance, both of the images shown in Fig.\ \ref{samevote_1}, which were displayed as a pair in StudyMB 1.0, were assigned the same quality score, although the quality differences between them become obvious when inspected in detail. 
Another limitation of StudyMB 1.0 is that in the paired comparison crowdworkers were asked to identify the image with the better quality and not the image which better matches with the ground truth image that was not shown. This may have created a bias as also ground truth images may carry some distortions, in particular motion artefacts.

The primary metric to evaluate the performance of objective IQA methods is the SROCC, i.e., a rank-order correlation. 
An increased accuracy of image quality ranking could benefit the development of objective IQA methods. Thus, we improved the design of the subjective study in the following three ways. (1) Artefact zooming: We help users identify the degradation by providing zoomed image portions that contain the most noticeable artefacts. (2) Artefact amplification: We increase the local differences of the interpolated images w.r.t.\ the ground truth images without significantly changing the average color properties of the images in the changed areas. (3) In the paired comparisons, we additionally provided the ground truth image, and the task was to identify the interpolated image closer to the ground truth. We will argue and show by an additional experiment that although our artefact zooming and amplification may change the perceived quality of an image, it will not distort the results of paired comparisons in terms of the ranking of the participating images.

In this paper, we implemented this improved paired comparisons of interpolated images given by optical flow algorithms in the Middlebury interpolation benchmark and re-ranked them accordingly (denoted as \emph{StudyMB 2.0}). Comparing the old ranking according to RMSE in the Middlebury benchmark and the re-ranking according to our improved subjective study then allows us to judge the suitability of existing quality metrics.

The outcome of our study is clear. It demonstrates that current FR-IQA methods are not suitable
for assessing the perceived visual quality of interpolated frames that have\linebreak been created by using optical flow algorithms. Consequently, using the collected subjective scores
as ground truth, we propose a novel FR-IQA method. It is based on a
weighted absolute error (WAE) which locally assigns different weights to absolute errors between the interpolated image and its ground truth. The average result of the leave-one-out (LOO) cross validation shows that WAE-IQA offers a slightly better performance for the interpolated Middlebury images than the currently best FR-IQA method from the literature.

Summarizing, compared to \citep{huiqomex2019}, the contribution of the current journal paper is threefold:

\begin{itemize}
    \item We provide better subjective quality scores via artefact amplification and zooming, which serve as a basis for the development and evaluation of new FR-IQA methods.  
    \item Based on the new scores, we reveal the poor perfor\-mance of existing FR-IQA methods when predicting the quality of motion compensated frame interpolation.
    \item We further propose a weighted error based FR-IQA me\-thod, which is specifically designed for the frame interpolation with motion compensation. 
\end{itemize}



\section{Related Work}
\vspace{-2pt}
As the recent literature on frame interpolation shows, there is mainly one benchmark that is considered for evaluating the performance of frame interpolation methods: the Middlebury benchmark. Originally designed for the evaluation of optical flow algorithms, this benchmark also offers an evaluation of motion-based frame interpolation results based on the calculated optical flow. To this end, it compares the interpolated frames with the ground truth in-between images that have been obtained by recording or rendering the original image sequence with a higher frame rate. Hence, despite of its original focus on evaluating optical flow methods, in the last few years this benchmark has become the de facto standard for
the evaluation and comparison of frame interpolation algorithms; see e.g. \citep{raket2012motion, meyer2015phase, niklaus2018context}.

Apart from the Middlebury benchmark, there are also two other datasets that are, however, less frequently considered for evaluation. While some interpolation algorithms like \citep{liu2017video, gongvideo} use the UCF 101 dataset \citep{soomro2012ucf1012} for training and testing, others like \citep{zhai2005low,ghutke2016novel} considered the videos from \citep{xidian1,xidian2}. 

Regarding the assessment of the interpolation quality, both the Middlebury benchmark and the other data sets rely on standard metrics, such as MSE, PSNR, or SSIM to measure the differences between the interpolated result and the ground truth in-between image. However, to the best of our knowledge, there have been no attempts so far to analyze how useful these metrics actually are to measure the quality of motion-compensated frame interpolation.


Regarding the amplification of artefacts we applied pixel value range expansion. This method was so far only adopted in reverse tone mapping operators (rTOMs), aiming to give low dynamic range images (LDR) the appearance of a higher dynamic range (HDR) without annoying artefacts \citep{wang2011low}. The general technique of rTOM first identifies the brightest areas of the image, yielding a certain kind of expansion map. Those bright areas are then expanded to a significant extent using different dynamic range expansion functions, whereas the rest of the pixels of the image are kept unchanged or only slightly modified \citep{masia2017dynamic}. To the best of our knowledge, such idea of amplifying the pixel value range for increasing the error visibility was not adopted in any subjective image quality assessment study so far.


\section{Prior Knowledge}
\subsection{Subjective Study}
\emph{Absolute Category Rating} (ACR) is a type of subjective testing where the test items are presented one at a time and are rated independently on one of five possible ordinal scales, i.e., Bad-1, Poor-2, Fair-3, Good-4, and Excellent-5 \citep{itu1999subjective}.

ACR is easy and fast to implement, however, it has several drawbacks \citep{chen2009crowdsourceable}. Participants may be confused when the categories of the rating scale have not been explained sufficiently well. They may also have different interpretations of the ACR scale, in particular in crowdsourcing experiments because of the wide range of cultural backgrounds and perceptual experiences of the crowd workers. 
Moreover, the perceptual distances between two consecutive scale values, e.g., between 1 and 2, should ideally be the same. However, in practice this can hardly be achieved \citep{hossfeld2016qoe}. Also it is not easy to detect when a participant intentionally or carelessly gives false ratings.
Alternatively, {\em paired comparisons} (PC) can solve some of the problems of ACR. In a PC test, items to be evaluated are presented as pairs. In a forced-choice setting, one of the items must be chosen as the preferred one. The main advantage of this strategy is that it is highly discriminatory, which is very relevant when test items have nearly the same quality.

However, when implemented naively, comparing $N$ items would require ${N \choose 2 }$ comparisons, too many to be practical, when $N$ is on the order of 100, for example. In our case, for each of the 8 sequences, we would have to compare $N=155$ images, giving a total of 95,480 pairs. 

A practical solution to this problem is to resort to the concept of randomly paired comparisons that is based on randomly choosing a fraction of all possible paired comparisons. This strategy is not only more efficient, it also has been proven
to be as reliable as full comparisons \citep{xu2012hodgerank}. After obtaining results from 
these comparisons, subjective scores have to be reconstructed. This can be done based on
Thurstone's model \citep{thurstone1927law,luce1994thurstone} or the Bradley-Terry model \citep{bradley1952rank}. 

\subsection{Thurstone's Model}\label{thurstone}
Thurstone's model provides the basis for a psychometric method for assigning scale values to options on a 1-D 
continuum from paired comparisons data. It assumes that an option's quality is a Gaussian random variable, thereby accommodating differing opinions about the quality of an option.
Then each option's latent quality score is revealed by the mean of the corresponding Gaussian. 

The result of a paired comparison experiment is a square count matrix $C$ denoting the number of times that each option was preferred over any other option. More specifically, for $n$ comparisons of option $A_i$ with option $A_j$, $C_{i,j}$ gives the number of times $A_i$ was preferred over $A_j$. Similarly, $C_{j,i}$ in the count matrix denotes the number of times that $A_j$ was preferred over $A_i$, and we have $C_{i,j}+ C_{j,i}=n$. 

According to Thurstone's Case V, subjective qualities about two options A and B are modelled as uncorrelated Gaussian random variables $A$ and $B$ with mean opinions $\mu_A$, $\mu_B$ and variances ${\sigma_A}^2,{\sigma_B}^2$, respectively. When individuals decide which of the two options is better, they draw realizations from their quality distributions, and then choose the option with higher quality. More specifically, they choose option A over option B if their sample from the random variable $A-B$ (with mean $\mu_{AB} = \mu_{A}  - \mu_{B}$ and variance ${\sigma_{AB}}^2 = {\sigma_{A}}^2 + {\sigma_{B}}^2$) is greater than $0$. Therefore, the probability of a subject to prefer option A over B is: 
\begin{ceqn}
\begin{align}
\label{eq1} 
   P(A > B) = P(A - B > 0)= \Phi \left( \frac{ \mu_{AB} }{ \sigma_{AB} } \right),
\end{align}
\end{ceqn}
where $\Phi(\cdot)$ is the standard normal cumulative distribution function (CDF). 

Thurstone proposed to estimate $P(A > B)$ by the empirical proportion of people preferring A over B, which can be derived from the count matrix $C$ as: 
\begin{ceqn}
\begin{align}
\label{countmat}
    P(A>B) \approx \frac{C_{A,B}}{C_{A,B}+C_{B,A}}.
    \end{align}
\end{ceqn}
The estimated quality difference $\hat{\mu}_{AB}$ can be derived from inverting Eq.\ \ref{eq1}, giving: 
\begin{ceqn}
\begin{align}
\hat{\mu}_{AB}=\sigma_{AB} \Phi ^ {-1} \left( \frac{C_{A,B}}{C_{A,B}+C_{B,A}} \right)
\end{align}
\end{ceqn}
known as Thurstone's Law of Comparative Judgment,  where $\Phi^{-1}(\cdot)$ is the inverse standard normal CDF, or z-score. Least-squares fitting or maximum likelihood estimation (MLE) can be then applied to estimate the scale values $\mu_{A}$ for all involved stimuli $A$. For more details we refer to \citep{tsukida2011analyze}.

\section{Subjective Quality Assessment: StudyMB 2.0}

\subsection{Data and Study Design}

In order to re-rank the methods in the Middlebury benchmark, we implemented paired comparisons based on Thurstone's model with least-squares estimation to obtain subjective judgments of the image qualities. In the benchmark, there are 8 sets of 155 interpolated images each, most of which had been generated by optical flow methods.\footnote{When we ran the experiments in March 2019, there were altogether 155 methods in the Middlebury benchmark, including a number of additional, new methods compared to \citep{huiqomex2019}}. To run a complete set of possible comparisons  would require collecting ratings for $8 \times 155 \times 154 / 2$ pairs, which is too many for practical purposes. However, it is sufficient to compare only a subset of these pairs. Therefore, we randomly sample pairs within each of the 8 sets, such that the 155 images form a random sparse graph with a vertex degree of 6, i.e., each image was to be randomly compared to 6 other images, which resulted in $ 465 \times 8 = 3720 $  pairs of images. We ran the experiment using the Amazon Mechanical Turk (AMT) platform \citep{amt}. We mixed the eight sets of image pairs and randomized their display order. Crowd workers were shown, in turn, a single pair of images per page. Payments were made for each completed pair. We collected 20 votes per pair. In total, 293 crowd workers participated in our experiment.

\begin{figure}[t]
\centering{\fbox{\includegraphics[width=0.48\textwidth]{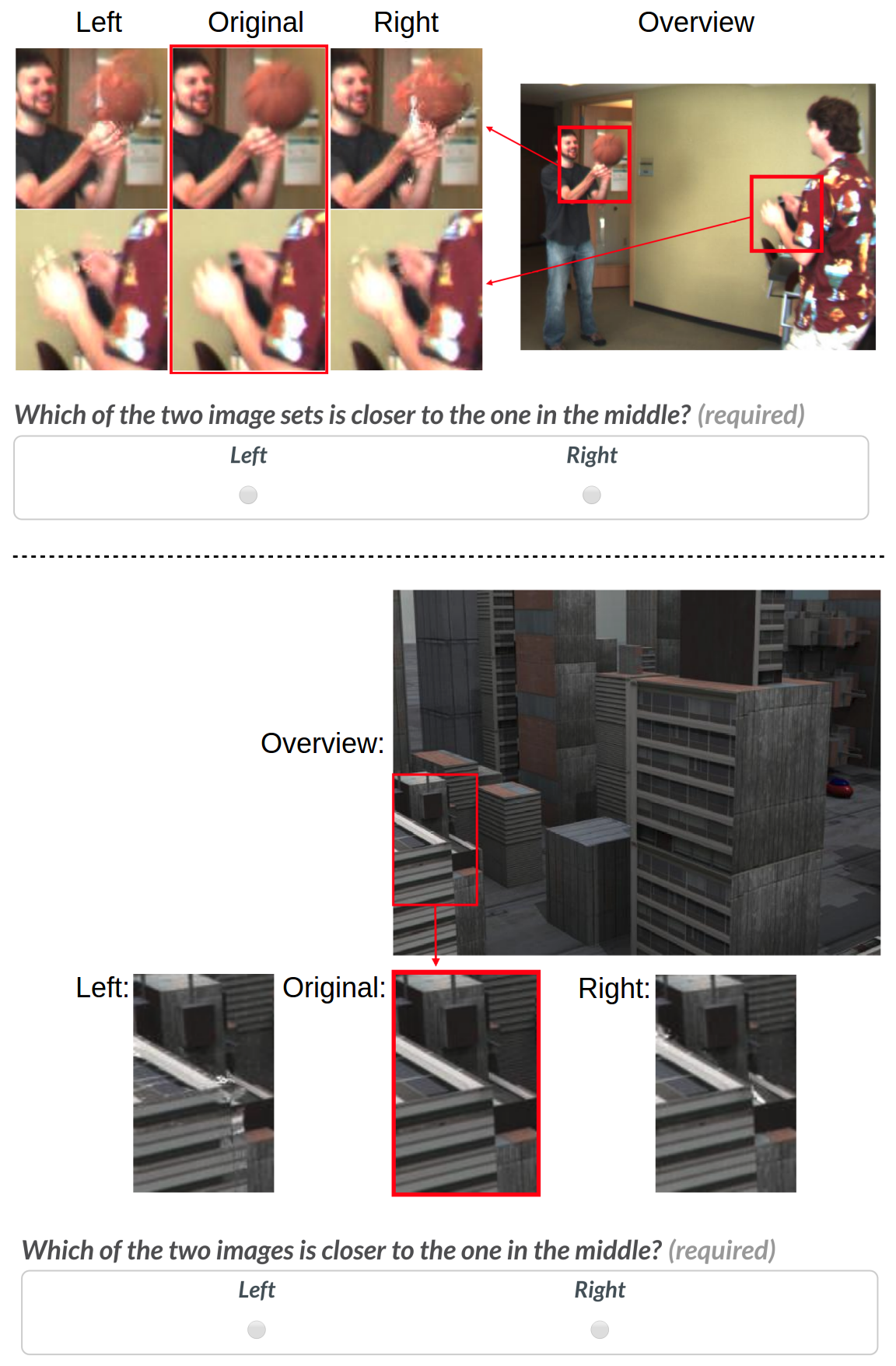}}}
\caption{Interface of the crowdsourcing experiment. The full-sized ground truth image was displayed in the experiment overview section, with the most noticeable degraded parts highlighted. Zoomed portions of the interpolated images were provided as pairs, with the ground truth placed in the middle.}
\label{inter}
\end{figure}

In order to increase the sensitivity of the subjective detection of minute differences between two interpolated images, we applied two methods.
\begin{enumerate}
    \item[(1)] \textbf{Artefact amplification.} Interpolated images in the benchmark differ from the ground truth images. The pixel-wise differences w.r.t.\ the ground truth images were artificially increased for display and judgment.
    \item[(2)] \textbf{Artefact zooming.} Artefacts due to interpolation based on optical flow tend to be localized in images, for example, nearby edges of moving objects. To steer the attention of the crowd workers towards these most heavily degraded image portions, these regions were displayed also enlarged below the full image above.
\end{enumerate}
The crowdsourcing interface for one comparison is shown in Fig.~\ref{inter}. It contains the zoomed image regions with the most severe distortions for each of the images to be compared and additionally the ground truth image, in full size and with the zoomed portion. In the next two subsections we give the details for these two methods.


\subsection{Boosted Paired Comparison} 

When evaluating the performance of objective IQA methods, the primary metric is the SROCC, i.e., a rank-order correlation. The absolute quality values are secondary; however, their ordering is most relevant. Thus, if the ground truth for an objective IQA would rank an item $A$ higher than $B$, then it is desired that this relation also holds true for the estimated qualities of $A$ and $B$ by the objective IQA method. If one directly compares the quality values between ground truth and predictions, one usually applies a nonlinear regression beforehand to align the predicted quality values as much as possible with the ground truth quality values. Only after such an alignment the Pearson linear correlation (PLCC) gives results that are suitable for comparison of different objective IQA methods.

Similarly, for the Middlebury benchmark it is the ranking of the different methods for optical flow, and respectively for frame interpolation that is most relevant and not the particular scalar quality values themselves. For the purposes of our research, therefore, the precision of the absolute values in our subjective quality study is less important than their correct perceptual ordering. The boosted paired comparison, i.e., the artefact amplification and the zooming basically is designed as a monotonic and even linear operation (spatial scaling by zoom and linear scaling of pixel intensity differences) which can be expected not to reverse the subjective ordering for perceived quality.

In order to clarify this effect, we have carried out an additional crowdsourcing experiment that demonstrates the increased sensitivity of boosted paired comparison. Moreover, it shows that the accuracy of the ranking of the resulting quality values is not only maintained w.r.t.\ the case without boosting, but even increased (see Subsection \ref{sec_JPEG}). Before we get to that, we present the technical details of the proposed artefact amplification and zooming in the following two subsections.

\subsubsection{Artefact Amplification by Local RGB Transformation} 

Since the human visual system is not sensitive enough to reliably detect the quality differences for many of the interpolated images \citep{huiqomex2019}, we artificially emphasized the defects by linearly scaling up the differences relative to the ground truth. Following Weber's law \citep{fechner1966elements}, given an interpolated image, we basically enlarged the RGB pixel differences for all pixels linearly, following
$$
    {\hat v}'=v+\alpha (\hat v-v),
$$
where $v$ is an original RGB (ground truth) component value, and $\alpha > 1$ is a fixed amplification factor. For 24 bit color images, the transformed RGB component values should not be outside the range $[0,255]$, which could be achieved by common clamping. However, clamping is a nonlinear operation which might actually remove differences in artefacts between two interpolated images. Therefore, we propose to reduce the amplification factor $\alpha$ at pixels where the linear RGB transformation would require clamping, to the maximal $\alpha$ without requiring clamping. See Algorithm \ref{alg1} for the details.


\begin{algorithm}[t]
\caption{Pixel-wise Artefact Amplification}
\label{alg1}
\begin{algorithmic}[1]
\State $\alpha  \gets$ default value \Comment{$\alpha \gets 2$ for the images in StudyMB 2.0}
\State $v \gets (v_\text{r},v_\text{g},v_\text{b})$ \Comment{ground truth image pixel}
\State $\hat v \gets (\hat{v}_\text{r},\hat{v}_\text{g},\hat{v}_\text{b})$ \Comment{interpolated image pixel}
\For{r component}
\If{$\hat{v}_\text{r}-v_\text{r} > 0$}
     \State $\alpha_{\text{r},\max} \gets (255-v_\text{r})/(\hat{v}_\text{r}-v_\text{r})$
\ElsIf{$\hat{v}_\text{r}-v_\text{r} < 0$} 
     \State $\alpha_{\text{r},\max} \gets -v_\text{r}/(\hat{v}_\text{r}-{v_\text{r}})$
\Else
      \State $\alpha_{\text{r},\max} \gets \alpha $
\EndIf
\EndFor
\For{g, b components}
\State same as above for r component
\EndFor
 
 \State $\alpha \gets \min(\alpha,\alpha_{\text{r},\max} , \alpha_{\text{g},\max}, \alpha_{\text{b},\max}) $
 \State $v' \gets v + \alpha (\hat{v} -v) .$
\end{algorithmic}
\end{algorithm}

\begin{figure*}[h!]
\centering{\includegraphics[width=1\textwidth]{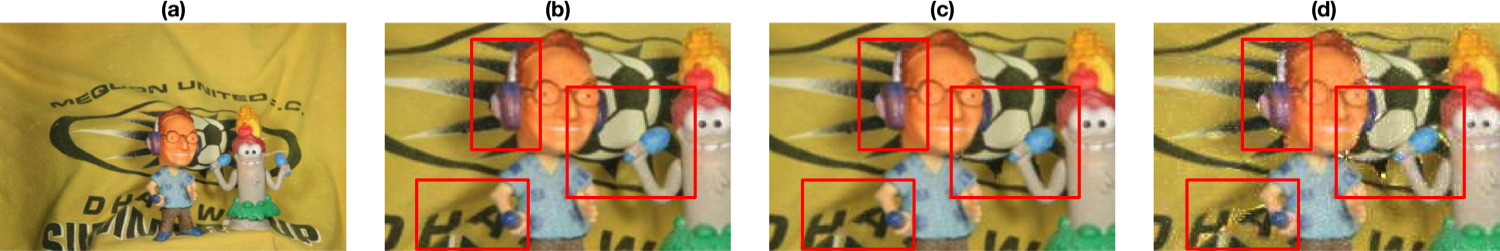}}
\vspace{-5pt}
\caption{(a) Full size ground truth image of \emph{Mequon}. (b) Zoomed part of (a). (c) Zoomed part of an interpolated image (original, without artefact amplification). (d) Artefact amplified version of (c). Distortions in (d) are more visible than in (c), especially for the parts in the red rectangles.}
\label{amplimequon}
\end{figure*}

\begin{figure*}[h!]
\centering{\includegraphics[width=1\textwidth]{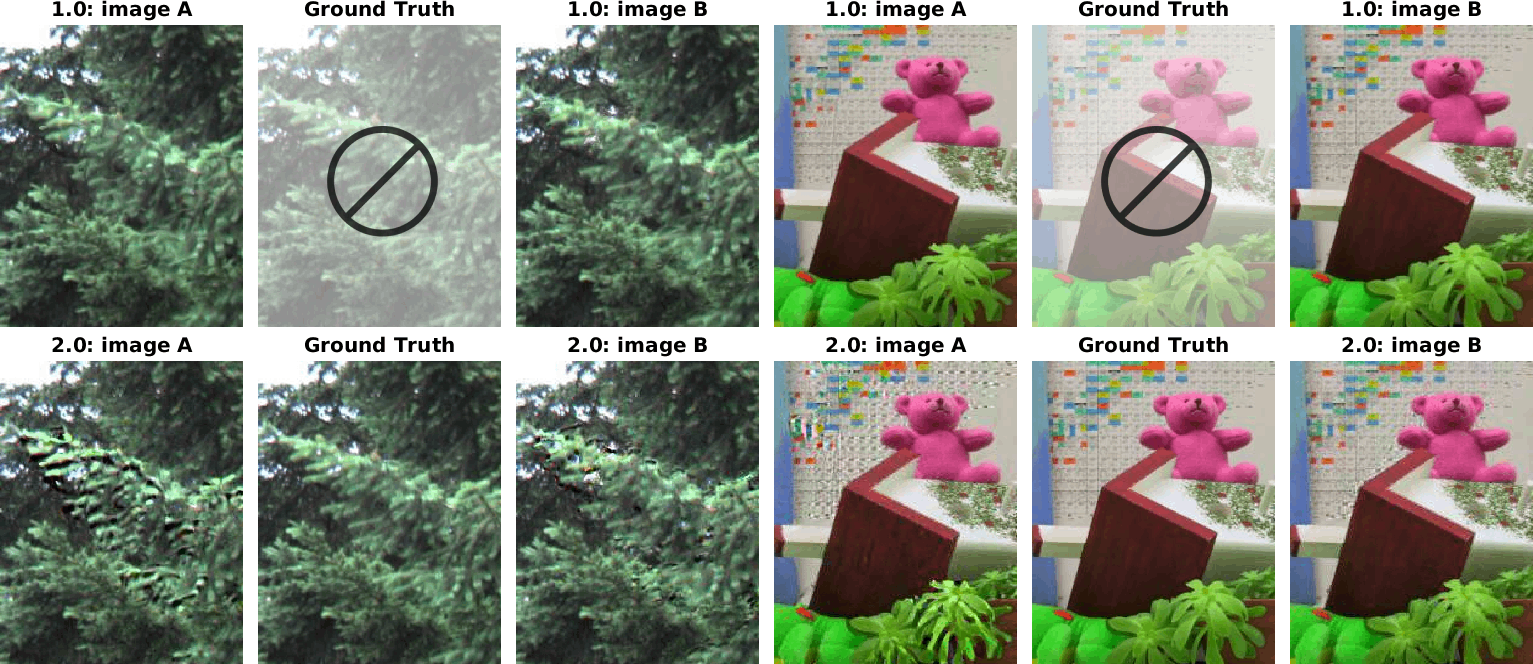}}
\caption{Top row: examples of degraded images ($A$ and $B$) used in StudyMB 1.0. Bottom row: the corresponding amplified versions from StudyMB 2.0, together with the ground truth images. The ground truth was not available to study participants in StudyMB 1.0. For the top row images in StudyMB 1.0, the quality differences are difficult to distinguish. After artefact amplification, the distortions become much more noticeable (second row), in particular when compared to the ground truth image. The distortions in images $A$ in both examples become more obvious, compared to those in images $B$. For instance, the amplified distortions in image $A$, left example, bottom row, cover a wider area and have a higher magnitude than those in image $B$. Similarly, for the example on the right, the distortions in image $A$ on the left of the teddy-bear, and the bottom leaves, are prominent, appearing to be over-sharpened.}
\label{fig:examples}
\end{figure*}

As shown in Fig.\ \ref{amplimequon} and \ref{fig:examples}, before the pixel value range amplification, it is extremely hard to distinguish the quality differences between the original pair of images. After the amplification the differences become much more obvious and thus easier for participants in our experiment to provide reliable annotations.

\subsubsection{Artefact Zooming}
\begin{figure}[h!]
\centering{\includegraphics[width=0.48\textwidth]{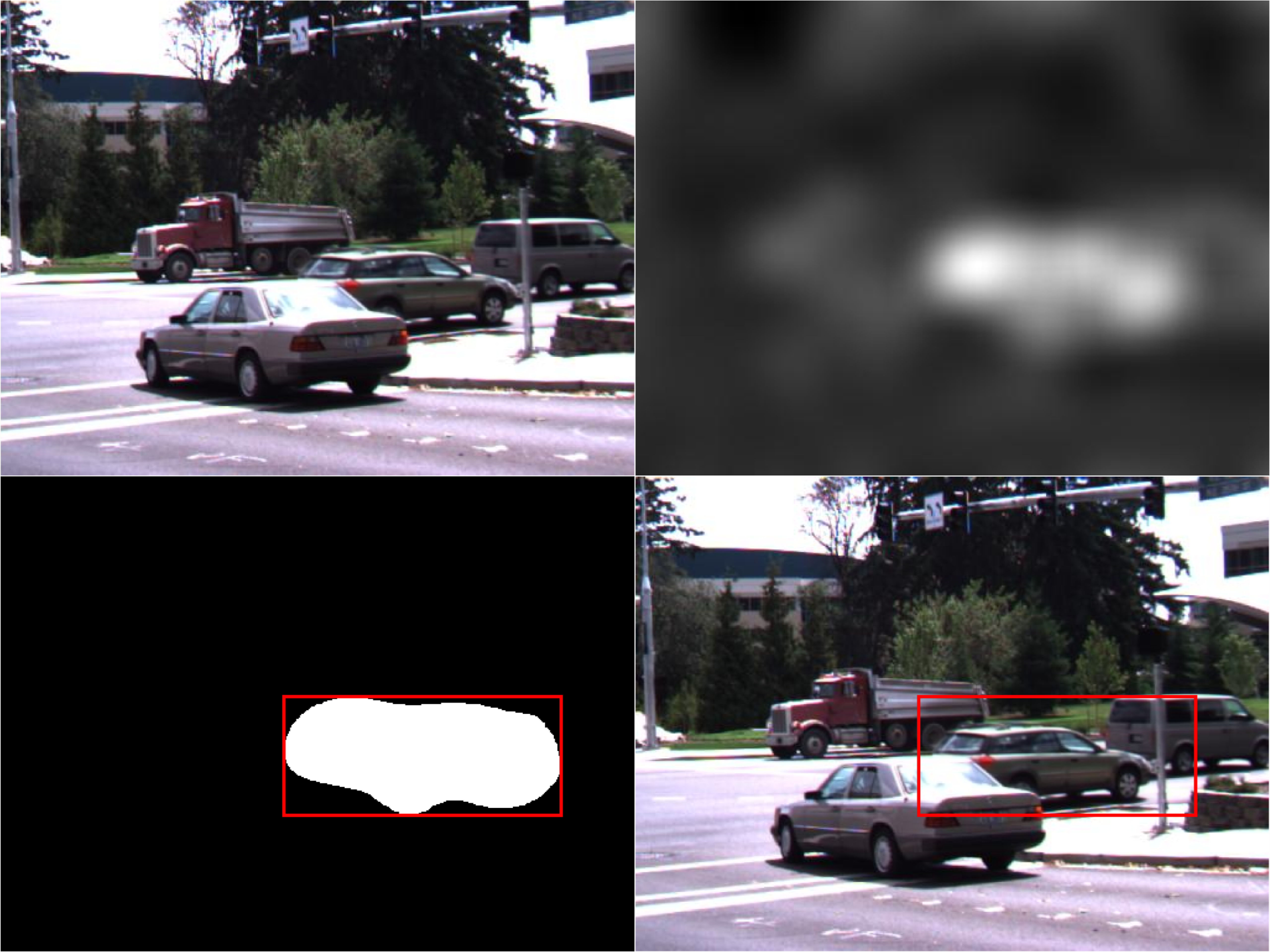} }
\vspace{-8pt}
\caption{Extraction of the most noticeable degraded portions of \emph{Dumptruck}. Upper left: average image (of 155 images in total). Upper right: Gaussian smoothed average error image ${\tilde{E}_{avg}}$. Lower left: segmentation of ${\tilde{E}_{avg}}$ using Otsu's method. Lower right: corresponding bounding box in the original image.}
\label{maindeg}
\end{figure}

In addition to boosting pixel value differences, we also\linebreak zoomed into the relevant regions that have the most noticeable degradation. Such regions of each set of images were extracted via the following steps (see Fig. \ref{maindeg}):

\begin{itemize}
    \item \emph{Step 1: Gaussian Smoothed Average Error Image}
    
       For the $n$-th image $I_n$ of 155 interpolated images from the same scene, the absolute error compared to the ground truth image $I_0$ was first computed, giving an absolute error image        $ E_n = | I_n - I_{0}|.$
       The mean of all 155 absolute error images is 
        $$E_{avg} = \frac{1}{155} \sum\limits_{n=1}^{155} | I_n - I_{0}|.$$
        
      A Gaussian filter (standard deviation of 20) was applied to the average error image $E_{avg}$, resulting in the Gaussian smoothed average error image, ${\tilde{E}_{avg}}$.
    
\item \emph{Step 2:  Segmentation}

Using Otsu's method \citep{otsu1979threshold}, the smoothed average error image ${\tilde{E}_{avg}}$ was segmented into two or more parts. The most noticeable  degraded portions were then extracted by bounding boxes around the segmented parts. 

\end{itemize}

\subsection{On the Potential of Boosted Paired Comparisons}
\label{sec_JPEG}
The purpose of boosted paired comparison is to increase the perceptual sensitivity w.r.t.\ small distortions allowing to better distinguish different distortion magnitudes. In order to demonstrate this effect, we need to change to an experimental context in which the ordering of ground truth quality is available.\footnote{
We cannot use the images from the Middlebury benchmark because there is no ground truth quality or ordering available.} For this purpose we have chosen a sequence of images, that were reconstructed from JPEG image compressions of the same pristine original image at different bitrates. Here we may assume that a higher bitrate corresponds to higher image quality, which provides the ground truth ordering of the image sequence.

In order to investigate the effect of artefact amplification and zooming, we carried out two PC experiments. The first, denoted as \emph{Plain PC}, is the conventional PC experiment. The second experiment carries out the same set of PCs, however, with artefact amplification and artefact zooming, denoted as \emph{Boosted PC}.

We took 12 JPEG distorted images, with a resolution of $1920\times 1080$ pixels (width$\times$height), at 12 distinct quality levels, from the MCL-JCI dataset \cite{jin2016statistical}. Let us denote them by $I_0,\ldots,I_{12}$, where $I_0$ is taken to be the original source image, and the ordering is with increasing distortion levels. We cropped the images to $600 \times 480 $ pixels for the plain PC experiment. For the boosted PC experiment, we manually cropped the images to a resolution of $280 \times 310$ pixels. After amplifying the artefacts using Algorithm \ref{alg1} with amplification factor $\alpha =1.5$, we spatially scaled them up by a zoom factor of 1.5, resulting in images of size $420 \times 465$. 

\begin{figure}[h!]
\centering{\includegraphics[width=0.48\textwidth]{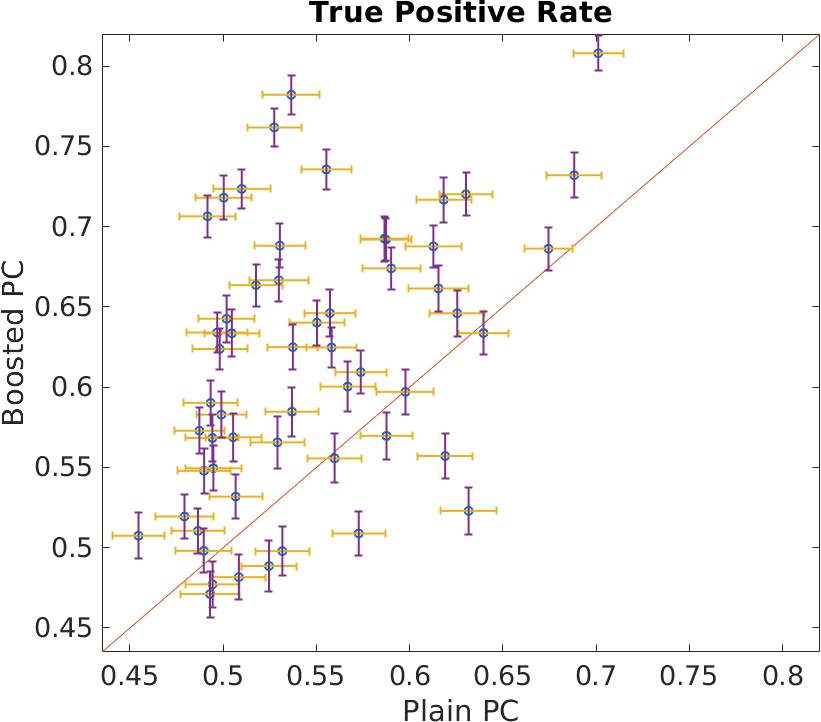} }
\caption{True positive rate (TPR) and 95\% confidence intervals for 57 image pairs (after 100 times bootstrapping).}
\label{tpr_pilot}
\end{figure}

For each experiment, we had the 57 image pairs $(I_k,I_l)$ with $|k-l| \le 6$ and $k < l$ judged by crowdsourcing on the AMT platform. For each PC, we collected 50 subjective ratings. We evaluated and compared the performances of the two experiments in two ways. 1) We computed the SROCC between the subjective quality scales and the ground truth ordering according to bitrate. 2) We computed the true positive rate (TPR) for each PC. The TPR is defined as the fraction of judgments that correctly indicated the better quality image, i.e., the image that was less distorted in a pair. Increasing the sensitivity of paired comparisons would result in an increase of the TPR. The results are shown in Fig.~\ref{tpr_pilot} and Table \ref{tb:srcc_pilot}. As can be seen, the TPR of boosted PC is much higher than that of plain PC in most cases. Moreover, boosted PC provided  a much higher rank order correlation (SROCC) to the ground truth as compared to plain PC. This indicates that boosted PC not only increases the sensitivity of the PC test, but also increases the accuracy of the ranking.\footnote{The above investigation was taken from a larger set of experiments on boosted PC for several types of distortions. That pilot study is included in the supplementary materials for further information of the reviewers.}

\begin{table}[t]
\caption{Rank order correlations and their confidence intervals between scale reconstructions and ground truth. The SROCC values shown are the means from 100 bootstraps.}
\label{tb:srcc_pilot}
\centering
\begin{tabular}{l |c c }
 & SROCC  & CI (95\%)  \\
 \hline
Plain PC & 0.8895 & [0.8771, 0.9018] \\
Boosted PC & 0.9864 & [0.9846, 0.9881]\\
\end{tabular}
\end{table}

\subsection{Outlier Removal}

We have little control over the experimental environment of crowd workers on AMT. In order to obtain high-quality experimental results, reliability controls are required. The de-facto control mechanism for reliability screening in crowdsourcing is test questions, regarding content and consistency \cite{hossfeld2013best}. However, we do not know the ground truth, nor do we have expert opinions to rely on as in \cite{Hosu2018-expertise-screening}. Using personal judgment to define test questions may wrongly bias user opinions towards the authors' expectations.
Thus, we decided not to use test questions during the experiment. Nonetheless, it is essential to screen workers after the crowdsourcing experiment is completed, in order to remove ill-intended or non-attentive participants, and improve scoring reliability. For this purpose, there are several established procedures that have been developed under the umbrella of user rating screening (URS) \cite{hossfeld2013best}. All ratings of identified unreliable workers should be removed from the analysis of the data. 

We propose an iterative URS method that is particularly suited for paired comparison experiments. In our case we have 8 sets of images that are scaled independently from PCs of many crowd workers. For simplicity, let us first restrict to a single set, and point out later how to generalize to several. 

The goal of the outlier detection is to identify a subset of workers whose collective PCs can be regarded as being in disagreement with the PCs of the remaining workers. To quantify the disagreement of a subset of workers with the rest, one may reconstruct the scale values from the remaining PCs as ``ground truth'' and then consider the true positive rate for the PCs of workers in the selected subset. Subsets with smaller TPR can be regarded as those exhibiting larger disagreement. In place of the TPR one can use the log-likelihood of the PCs of the subset of workers, which becomes available by means of the scale reconstruction and its underlying probabilistic model. This was done in \cite{perez2017practical} for all subsets consisting of a single worker, and workers were classified as outliers, based on a statistic of the resulting log-likelihood, followed by visual inspection.

In our approach we set a certain percentage of PCs that should remain and strive to identify a minimal subset of workers that most strongly disagrees with the remaining workers and such that the number of all retained PCs is close to the target percentage. Since it is computationally infeasible to proceed by complete enumeration, considering all subsets of workers, we propose an iterative method, as described in Algorithm \ref{alg:outlier}. Starting out with the reconstructed scale values, derived from the entire dataset of PCs, we compute the TPR for the set of PCs of each worker. We iteratively remove workers with the smallest TPR until the desired number of retained PCs from the remaining workers is reached. From these remaining PCs we reconstruct updated scale values, and repeat the procedure, starting out again with the complete pool of workers. The iteration terminates successfully when the same set of workers is identified in two subsequent iteration steps. In the case with our data, just a few iterations sufficed.

The procedure generalizes straightforwardly to our case in which we have several sets of images which comparisons only between image pairs chosen from the same set. In each iteration, separate reconstructions, one per set, have to be made. For our experimental crowdsourcing data, we arrived at the resulting threshold on the worker TPR of 0.6209, keeping 40\% of the total set of PCs. Note that a TPR of 0.5 can already be achieved solely by guessing. The outlier iteration procedure converged after 4 iterations.


\begin{algorithm}[t]
\caption{Iterative Outlier Removal}
\label{alg:outlier}
\begin{algorithmic}[1]
\State $D \gets$ dataset of all forced choice paired comparisons (PCs)
\State $P \gets$ target fraction of PCs to be retained in $D$, $0<P<1$
\State $S' \gets$ reconstructed scale values from the full dataset $D$
\Repeat
    \State $S \gets S'$   \Comment{Update scale reconstruction result}
    \State For each worker compute average TPR w.r.t.\ $S$
    \State Sort workers according to increasing average TPR
    \State Iteratively remove from $D$ the PCs of workers with least TPR
    \State ... until a subset $D' \subset D$ of size $|D'| \le P\cdot|D|$ is obtained
    \State $S' \gets$ reconstructed scale values from the reduced dataset $D'$
\Until $S = S'$ or timeout
\State \Return Subset $D'\subset D$
\end{algorithmic}
\end{algorithm}

\section{Re-ranking Results} \label{result}

In our study we chose to compare image pairs only within each of the eight sets of 155 images. Using the collected comparative judgments we reconstructed absolute quality scale values for each image using Thurstone's model and the code provided by \citep{li2018hybrid}. However, because we did not have cross-set comparisons available, the range of values reconstructed for each set are independent of each other. We propose a simple procedure to align the reconstructed scales, by introducing virtual anchors. We added two fictitious images as anchors. One of them represents the worst quality among all the images, and the other one is like the ground truth image, with a quality that is better than that of all the other images. 

\begin{figure}[t]
\centering{\includegraphics[width=0.48\textwidth]{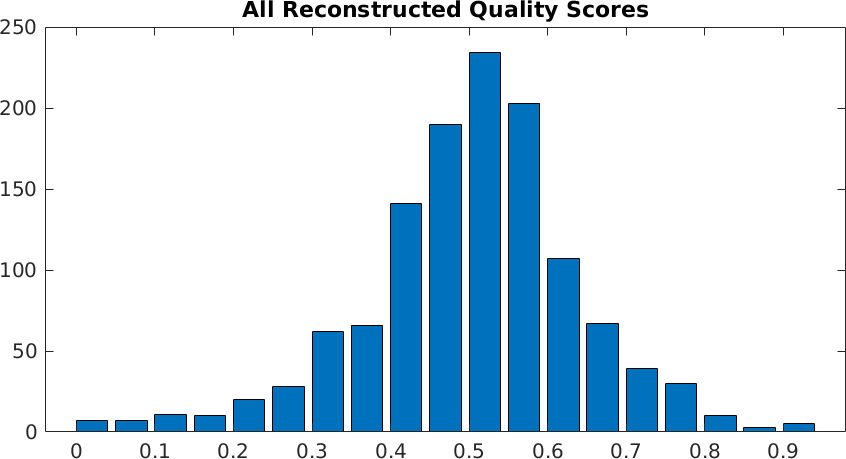}}
\vspace{-5pt}
\caption{Histogram of reconstructed quality scores of all the 8 sets of images obtained in StudyMB 2.0.}
\label{fig:allscore}
\end{figure}


\begin{figure}[t]
\centering{\includegraphics[width=0.48\textwidth]{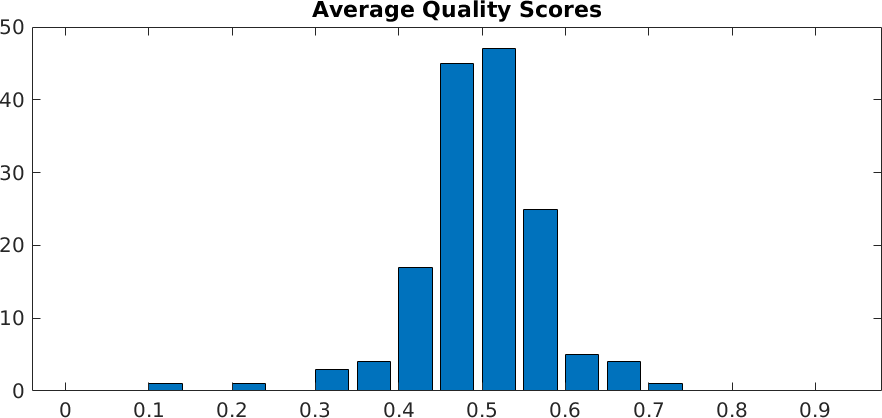}}
\vspace{-5pt}
\caption{Histogram of average quality scores over 8 sets obtained in StudyMB 2.0.}
\label{fig:avgscore}
\end{figure}

After reconstruction of the scale values for the 155+2 images in each set, we linearly rescaled the quality scores such that the quality of the virtual worst quality image became 0, and that of the ground truth image became 1. In this way, we rescaled the reconstructed scores to the interval $[0,1]$.\footnote{All images and their reconstructed quality values, accompanied by their corresponding rankings, are available on our website: http://database.mmsp-kn.de/. The differences between the re-ranking (ranked according to subjective study) and their corresponding ranking in the Middlebury benchmark (ranked according to RMSE) are also available on our website: http://database.mmsp-kn.de/.}

Each set was scaled and ranked separately. Then the average quality of a method was obtained by taking the mean of the (scaled) quality values of the 8 sets, which resulted in an overall rank. Fig. \ref{fig:allscore} shows the histogram of the reconstructed quality scores of all the 8 sets of images. Besides, the histogram of the average quality scores of 155 methods is depicted in Fig. \ref{fig:avgscore}.

\begin{figure}[t]
\centering{\includegraphics[width=0.48\textwidth]{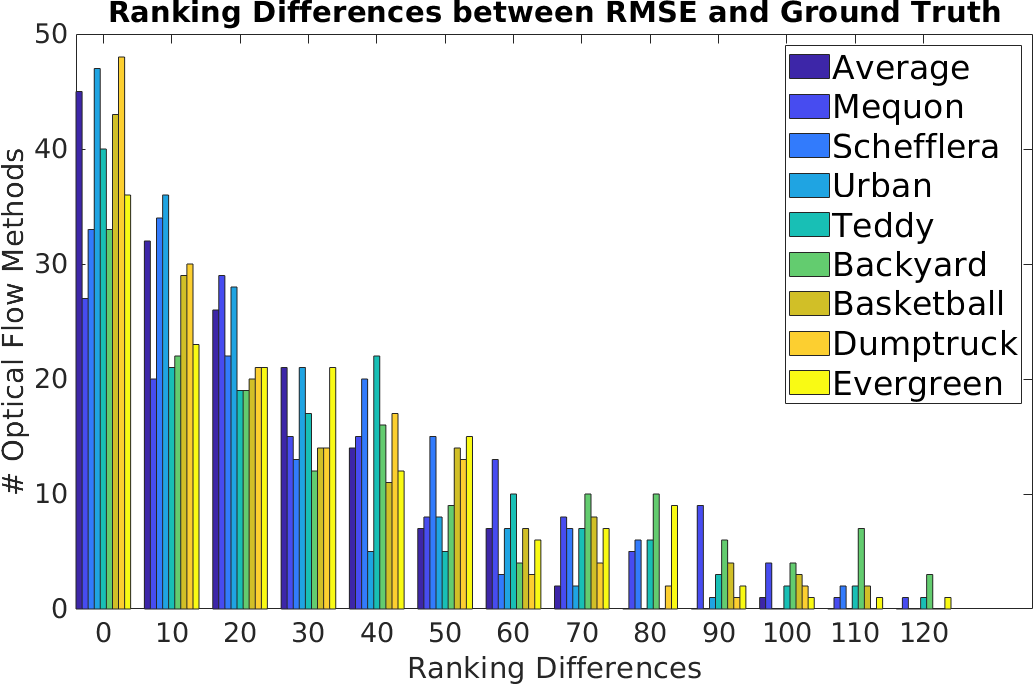}}
\vspace{-8pt}
\caption{Ranking differences between RMSE and ground truth (by StudyMB 2.0).}
\label{rankdiff_crdrmse}
\end{figure}

The best three methods ranked by the subjective study, i.e., SuperSlomo \citep{jiang2018super}, CyclicGen \citep{liu2019deep}, and CtxSyn \citep{niklaus2018context} ranked 3rd, 2nd, and 1st in the Middlebury benchmark, respectively. Overall, as shown in Fig. \ref{rankdiff_crdrmse}, 45 methods showed (average) rank differences within 10, and 31 methods gave (average) differences of more than 50 between their re-rankings and the rankings in the Middlebury benchmark.\footnote{Some methods are specifically tailored for frame interpolation (e.g., SuperSlomo) and others for optical flow estimation (e.g., DeepFlow2).  }

\begin{figure}[t]
\centering{\includegraphics[width=0.48\textwidth]{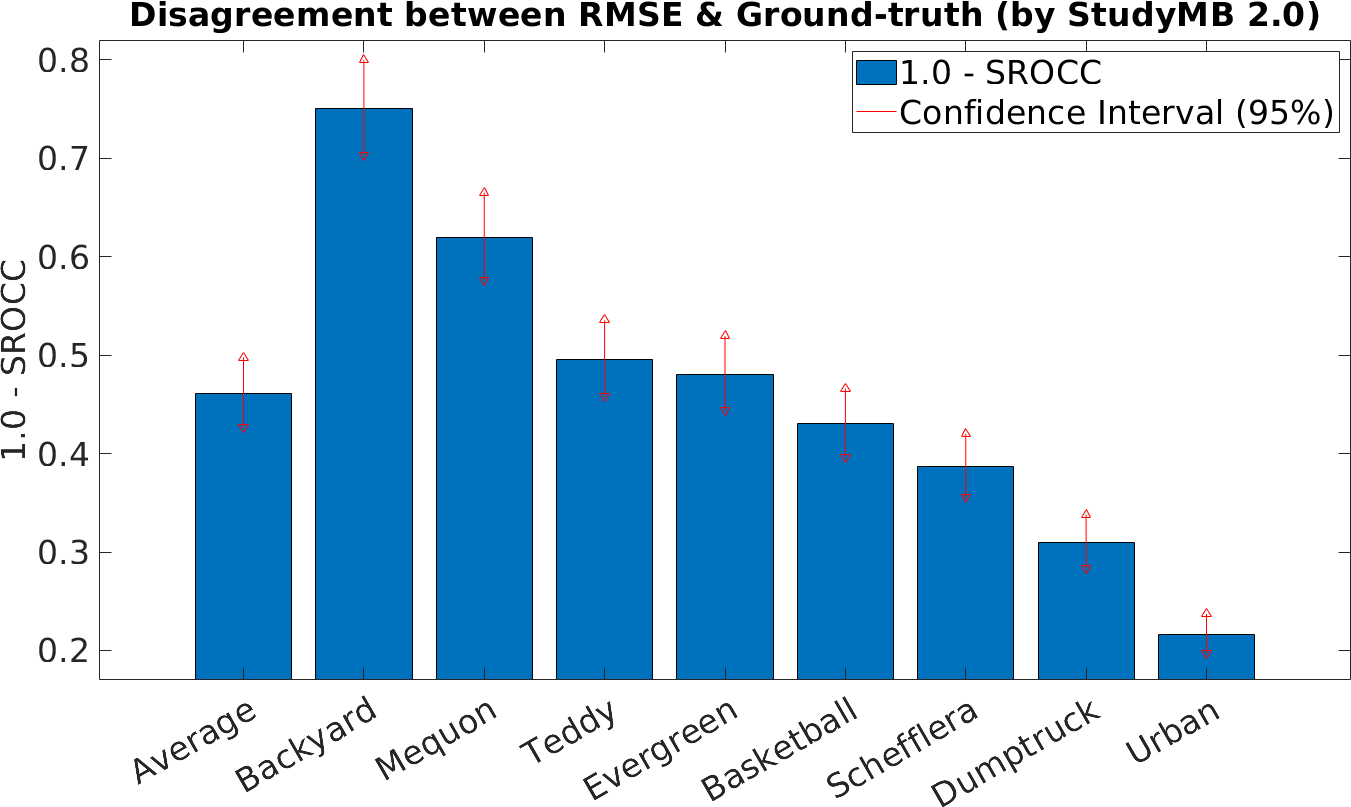}}
\vspace{-5pt}
\caption{Disagreement level with 95\% confidence interval.}
\label{visual}
\end{figure}

As an overall analysis, Table~\ref{tb:corrbootstrp} shows the bootstrapped (after 1000 iterations) SROCC values accompanied by the 95\% confidence intervals (CI) between the ranking in the Middlebury benchmark (i.e., ranking according to RMSE) and the re-ranking according to our subjective study. Note that the CI of SROCC was computed by transforming the rank correlation score into an approximate z-score using the Fisher transform \citep{ruscio2008constructing}. In a nutshell, a CI of probability $p$ is given by $\tanh(\arctan r \pm \Phi^{-1}(p) /\sqrt{n-3})$, where $r$ denotes the estimated SROCC, $n$ is the sampling size, and $\Phi^{-1}$ is the inverse of the standard normal CDF. In order to visualize the result, we computed the disagreement level as $1 - \text{SROCC}$ as shown in Fig.\ \ref{visual}.


\begin{table*}[h]
\caption{Correlations between RMSE and ground truth (by StudyMB 2.0) after 1000 iterations.}
\label{tb:corrbootstrp}
\centering
\resizebox{1\textwidth}{!}{
\begin{tabular}{c |c c c c c c c c c c c }
RMSE& Average & Mequon & Schefflera & Urban & Teddy & Backyard & Basketball & Dumptruck & Evergreen\\ 
\hline
SROCC	& 0.5389 & 0.3803 & 0.6135 & 0.7839 & 0.5043 & 0.2495 & 0.5699 & 0.6902 & 0.5193  \\
       
CI (95\%)	& [0.5025,0.5734] & [0.3352, 0.4237] & [0.5801, 0.6448] & [0.7630, 0.8031] & [0.4647, 0.5419] & [0.2004, 0.2973] & [0.5339. 0.6038] & [0.6621, 0.7164] & [0.4805, 0.5561]\\
\end{tabular}
}
\end{table*}

\begin{table*}[h]
\caption{Correlations of reconstructed MOS between StudyMB 1.0 and 2.0}
\centering
\scalebox{0.95}{
\begin{tabular}{l |c c c c c c c c c}
& Average & Mequon & Schefflera & Urban & Teddy & Backyard & Basketball & Dumptruck & Evergreen \\
\hline
PLCC & 0.5464 & 0.4453 & 0.2548 & 0.8358 & 0.4219 & 0.4287 & 0.8446 & 0.7399 & 0.4006 \\
SROCC & 0.5361 & 0.3883 & 0.2861 & 0.8322 & 0.4056 & 0.3977 & 0.8594 & 0.6909 & 0.4287 \\
KROCC & 0.3882 & 0.2711 & 0.1807 & 0.6387 & 0.2719 & 0.2699 & 0.6740 & 0.5086 & 0.2906 \\
 \end{tabular}
}
 \label{corr}
\end{table*}

\section{Comparisons between StudyMB 2.0 and 1.0}
As mentioned before, we implemented StudyMB 1.0 in our preliminary work \citep{huiqomex2019}. Similar as StudyMB 2.0, in StudyMB 1.0, we also used paired comparisons and crowdsourcing to collect subjective quality scores of the interpolated images in the Middlebury benchmark. However, there exist a number of differences between these two experiments, leading to differences in the subjective scores obtained.

\subsection{Differences in Data}
In StudyMB 1.0, we evaluated the interpolation performances of 141 optical flow methods. 14 more methods were added in StudyMB 2.0, resulting in 155 optical flow methods. Note that for later comparisons between these two experiments, we ignored those 14 more methods in StudyMB 2.0 to make them comparable. 


\subsection{Differences in Study Design}
The first difference regarding study design is that, instead of launching eight separate jobs (in StudyMB 1.0), each of which consists of the same contents, we mixed all the eight sets as one job in StudyMB 2.0. This avoided the phenomenon of semantic satiation \citep{das2014verbal,prochwicz2010semantic} which can occur not only in text but also in images. 

     
    

Secondly, in the instructions of StudyMB 2.0, we did not only highlight degraded parts as in StudyMB 1.0, but we also amplified the pixel value range of the interpolated images for increasing the visibility of the quality differences and zoomed the most noticeable parts, in order to draw attention to the regions of interest and make it easier for participants to perform their task.

The third difference is that in StudyMB 2.0, ground truth image was provided between two images that to be compared as a reference. Subjects were asked to select the one that can represent the ground truth more faithfully. While in StudyMB 1.0, no ground truth was provided. The subjects were requested to select the image of better quality.

\begin{figure}[h!]
\centering{\includegraphics[width=0.48\textwidth]{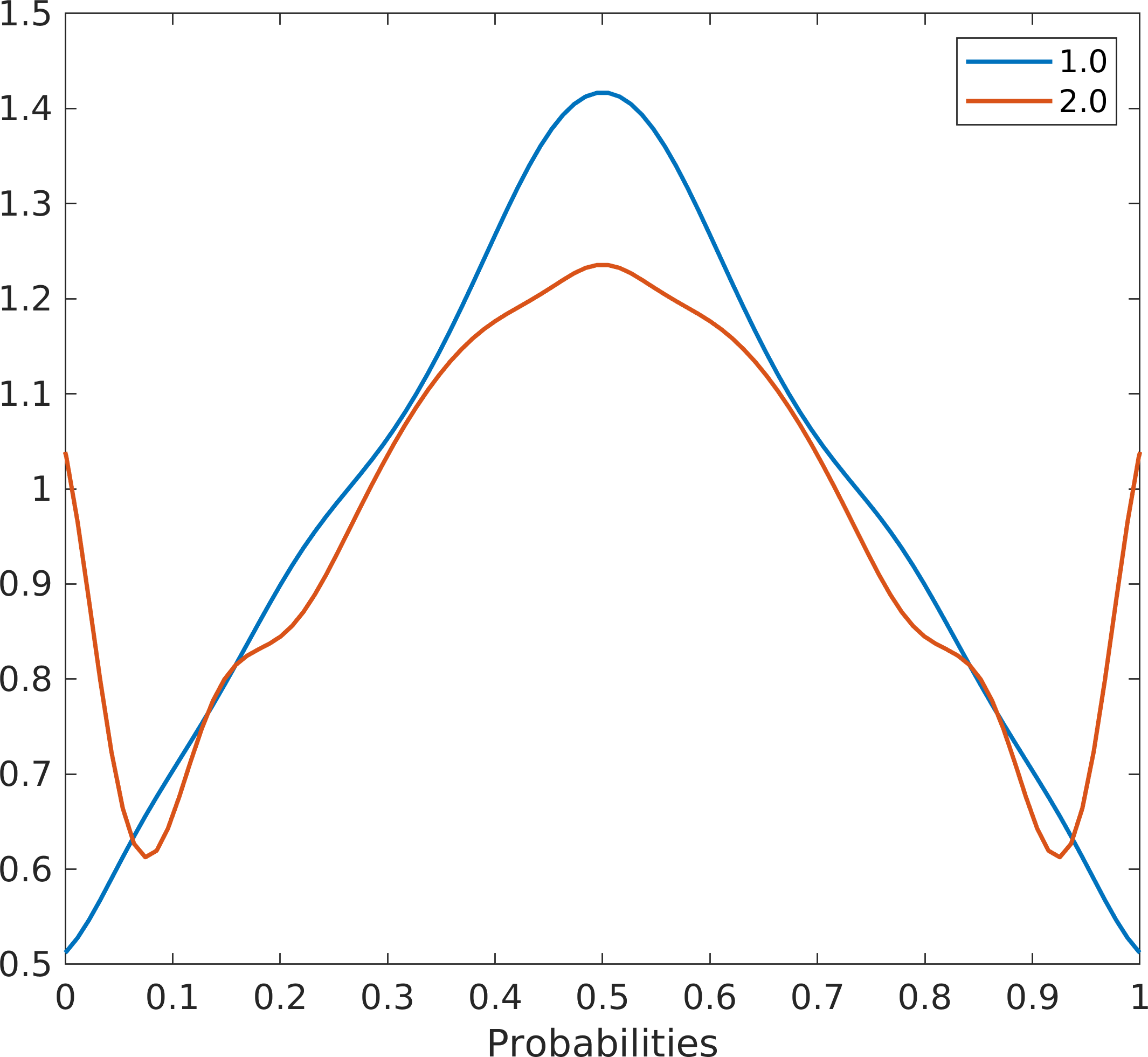}}
\caption{Density of aggregated probabilities in all the eight count matrices of StudyMB 1.0 and 2.0.}
\label{proball}
\end{figure}

\begin{figure}[t]
\centering{\includegraphics[width=0.48\textwidth]{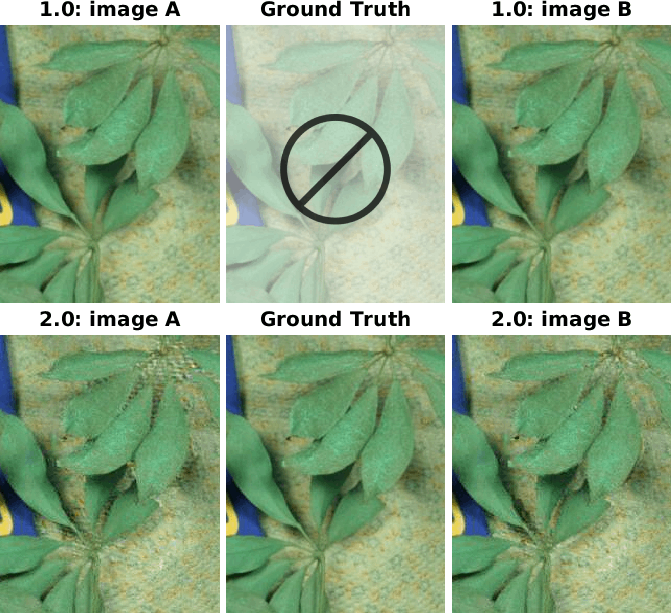}}
\vspace{-5pt}
\caption{Top row: example degraded images ($A$ and $B$) used in StudyMB 1.0, ground truth was not provided. Subjects were asked to select the image of better quality. Bottom row: the corresponding amplified versions from StudyMB 2.0 with the ground truth presented in between $A$ and $B$. Subjects were asked to compare images $A$ and $B$ with the ground truth to find which is more faithfully reproducing it. In StudyMB 1.0, image $A$ received a higher score than $B$, while in StudyMB 2.0, $A$ received a much lower score than $B$. In StudyMB 1.0, image $A$ looks smoother than image $B$. As no ground truth was provided as a reference, image $A$ was considered as of better quality. The smoothing effect was reduced due to artefact amplification, and the underlying degradation was emphasized. For the images in StudyMB 2.0, distortions are more obvious in image $A$ than in image $B$ when compared with the ground truth. Hence, image $A$ was given a lower quality.}
\label{fig:sch}
\end{figure}

\subsubsection{Sensitivity to Quality Differences}

As described in Equation \ref{countmat}, the probability $P(A>B)$ in the count matrix $C$ denotes the empirical proportion of people preferring A over B. Thus, $P=0.5$ illustrates that options $A$ and $B$ got exactly the same number of votes in the  subjective study (i.e., $A$ and $B$ are of the same quality), whereas $P=0$ or $P=1$ depicts the fact that either $A$ or $B$ obtained all the votes whereas the other one got no vote (i.e., $A$ or $B$ is much better than the other one). Therefore, the distribution of the probabilities in the count matrices can properly reveal the sensitivity to quality differences of the subjective study. To this end, we computed the cumulative distribution function (CDF) of the aggregated probabilities in all the count matrices (of all the eight sets) for both of the two experiments, and fitted their CDFs to probability density function (PDF). As shown in Fig. \ref{proball}, the PDF for StudyMB 1.0 has a triangular shape with a peak at probability 0.5, which illustrates that for many of the image pairs in the experiment, their quality differences are hardly distinguishable. 
In contrast, the peak at probability 0.5 was less pronounced StudyMB 2.0, which indicates that in StudyMB 2.0 differences of image quality were detected more frequently. Moreover,  the PDF for StudMB 2.0 has a higher density near probability 0 and 1 compared to StudyMB 1.0. This shows that in StudyMB 2.0 there were more image pairs for which the participants unanimously voted for for the left or right image.
Comparing the PDFs  between these two experiments shows that the study design of StudyMB 2.0 increased the sensitivity to image quality differences for the subjects as compared with StudyMB 1.0.


\subsubsection{True Ranking Fidelity}


In studyMB 1.0, the subjects were asked to select the better quality image without the ground truth being available as a reference. As shown in Fig. \ref{fig:sch}, this design has a negative effect. Some types of distortions, such as blurriness, in some cases, can enhance the perceptual quality, thus leading to a reversed ordering of the underlying techniques. This effect was avoided by design in StudyMB 2.0; the ground truth image was shown as a reference and subjects were requested to select the artefact-amplified image that more faithfully represents the ground truth.

\begin{figure}[t]
\centering{\includegraphics[width=0.48\textwidth]{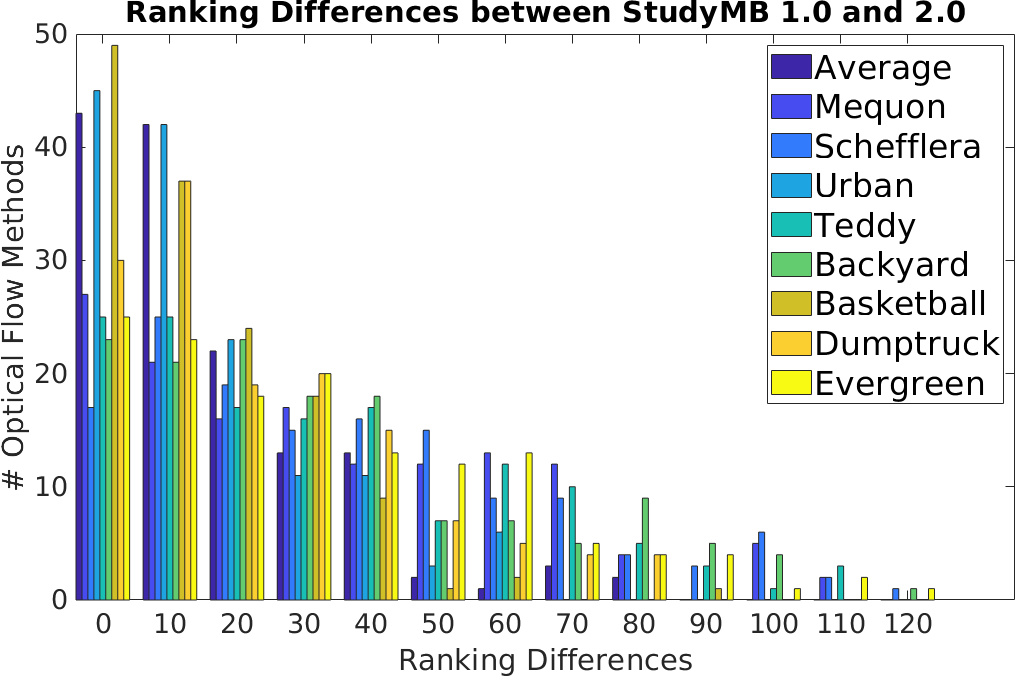}}
\caption{Re-ranking differences between StudyMB 1.0 and 2.0.}
\label{rankdiff}
\end{figure}

\subsection{Differences in Subjective Scores}
The differences in data and study design between StudyMB 1.0 and StudyMB 2.0 lead to the differences of subjective scores obtained for the interpolated images. In this regard, we investigated the correlations of MOS and re-ran\-king results between these two experiments.

\subsubsection{MOS Correlations}
We computed PLCC, SROCC, and Kendall's rank-order \linebreak  correlation coefficient (KROCC) between the subjective scores obtained in StudyMB 1.0 and 2.0. As shown in Table~\ref{corr}, the aforementioned three correlations between these two experiments are 0.55 (PLCC), 0.54 (SROCC) and 0.39 (KROCC) on average, which indicates an obvious difference of the subjective quality scores obtained for the same images.

\subsubsection{Re-ranking Differences}

Fig.~\ref{rankdiff} shows the differences of the re-rankings of 141 optical flow methods between StudyMB 1.0 and 2.0. It can be seen that, on average, there are {43 methods whose re-rankings differ fewer than 10 places, and 56 methods have re-ranking differences more than 20 places.} Overall, the re-ranking differences  further illustrate that there exist large differences of the quality scores obtained for the same interpolated images between these two experiments.

In summary, the main difference between StudyMB 1.0 and 2.0 is that we adopted artefact amplification as well as artefact zooming in StudyMB 2.0, which increased the sensitivity of the subjective study to a large extent. This gives rise to more precise and better scaled subjective quality scores of the interpolated images than the ones obtained in StudyMB 1.0.

\section{Weighted Absolute Error as FR-IQA}\label{sec:wae}

\begin{table*}[h]
\caption{SROCC of Rankings between FR-IQA and Our Subjective Study}
\centering
\scalebox{0.95}{
\begin{tabular}{l | c c c c c c c c c}
FR-IQA & Average & Mequon & Schefflera & Urban & Teddy & Backyard & Basketball & Dumptruck & Evergreen \\
\hline
MS-SSIM  & 0.5485 & 0.4517 & \textbf{0.6575} & 0.7249 & 0.4502 & 0.2527 & 0.7105 & 0.6274 & 0.5129 \\
RMSE & 0.5423 & 0.3831 & 0.6179 & 0.7868 & 0.5059 & 0.2531 & 0.5734 & 0.6930 & 0.5249 \\
VSI & 0.5397 & 0.4295  & 0.5166 & 0.7577 & 0.4810 & 0.2673 & 0.6568 & 0.6903 & 0.5180 \\
GMSD & 0.5383 & 0.4407 & 0.6038 & 0.7619 & 0.4605 & 0.2722 & 0.6087 & 0.6276 & 0.5313 \\
SSIM  & 0.5370 & \textbf{0.4715} & 0.6121 & 0.7067 & 0.4162 & 0.2383 & \textbf{0.7169} & 0.6281 & 0.5066 \\
FSIM  & 0.5339 & 0.4591 & 0.5732 & 0.7240 & 0.5088 & 0.2368 & 0.6009 & 0.6576 & 0.5106 \\
GN-RMSE & 0.5236 & 0.4515 & 0.6148 & 0.6868 & 0.3620 & 0.2719 & 0.7055 & 0.6147 & 0.4812\\
MAD  & 0.5220 & 0.3373 & 0.6256 & 0.6997 & \textbf{0.5236} & 0.2333 & 0.6009 & 0.6586 & 0.4970 \\
VIF  & 0.5133 & 0.4067 & 0.6032 & 0.7134 & 0.5130 & \textbf{0.4176} & 0.5065 & 0.4286 & 0.5171 \\
\hline
WAE-IQA & \textbf{0.5493} &  0.3910 & 0.6206 & \textbf{0.7969} & 0.5092 & 0.2703 &  0.5790 & \textbf{0.6947} & \textbf{0.5326}\\

 \end{tabular}
 }
\label{tb:crd-iqa}
\end{table*}

As described in Section \ref{result}, the re-ranking result of our subjective study shows that RMSE cannot reveal the perceptual quality of interpolated frames well. In fact, besides RMSE, most of the current FR-IQA methods cannot cope with the specific distortions 
that arise in interpolated frames produced by optical flow algorithms. As shown in Table~\ref{tb:crd-iqa}, seven of the most popular objective FR-IQA methods, SSIM \citep{wang2004image}, MS-SSIM \citep{wang2003multiscale}, MAD \citep{larson2010most}, FSIM \citep{zhang2011fsim}, GMSD \citep{xue2014gradient}, VSI \citep{zhang2014vsi}, VIF \citep{sheikh2004image} and GN-RMSE \citep{baker2011database}, gave low correlations with the subjective judgements. 

\begin{figure*}[h!]
\centering{\includegraphics[width=0.978\textwidth]{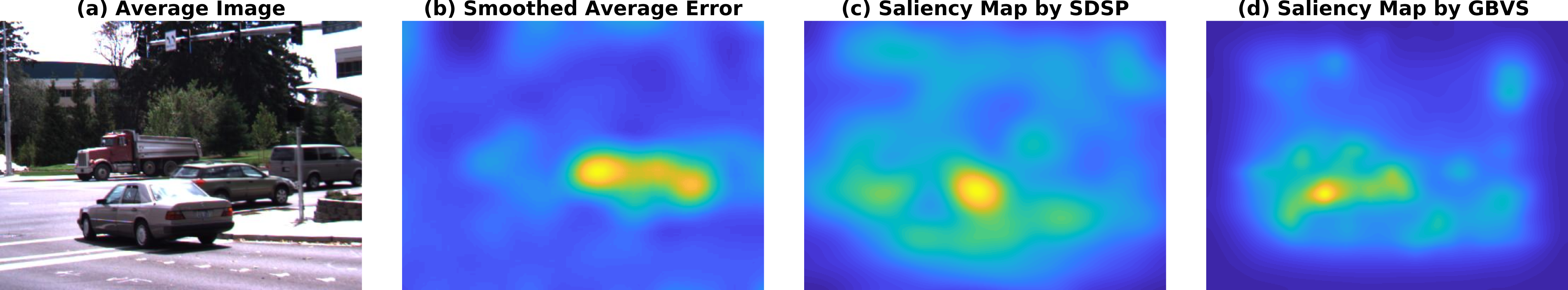}}
\caption{(a) Average image over 155 interpolated images of \emph{Dumptruck}. (b) Gaussian smoothed average error image (obtained by first taking the average error image over 155 images and then smoothed), which depicts the most degraded parts of image (a) compared to its ground truth. (c) Saliency map of (a) extracted using SDSP \citep{zhang2013sdsp}, the method adopted by VSI. (d) Saliency map of (a) extracted by GBVS.}
\label{vsi}
\end{figure*}

For example, VSI, one of the best FR-IQA methods based on saliency, yielded an SROCC of 0.952 when trained and tested on the LIVE database. 
However, when we applied the same method to the interpolated images by optical flow algorithms, VSI gave an SROCC of only 0.5397. This is likely due to two reasons:
\begin{itemize}
    \item The artefacts induced by optical flow algorithms lead to interpolated images that exhibit different, task-specific distortions that are not sufficiently taken into account in the IQA method.
    \item Saliency based methods like VSI focus on the most salient image regions. However, these may be just those that are not the most severely distorted parts, as shown in Fig.\ \ref{vsi}b and c.
\end{itemize}

Furthermore, we extracted the saliency map for the same image using GBVS \citep{harel2007graph}, one of the most widely used saliency detection methods. It can be seen from Fig.\ \ref{vsi}d that the resulting saliency map differs from the smoothed average error image to a large extent as well.

\begin{figure*}[h!]
\centering{\includegraphics[width=1\textwidth]{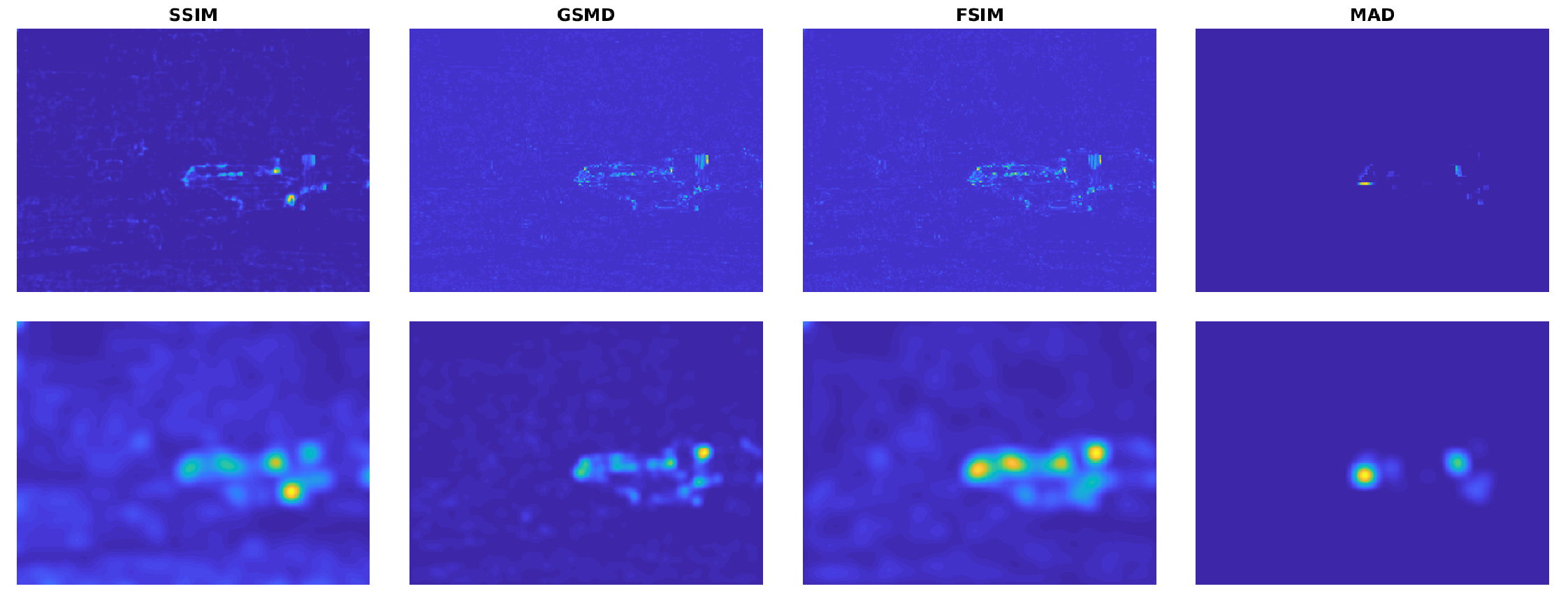}}
\vspace{-10pt}
\caption{Visualization of SSIM, GMSD, FSIM, MAD when computing the image quality between the average interpolated image (denoted as \emph{distorted image}) and its ground truth of \emph{Dumptruck}. \textbf{Upper row, first three:} distortion maps of SSIM, GMSD and FSI. The brighter the parts are, the more the distorted image 
differs from its ground truth. \textbf{Upper row, the last:} the estimation of the visibility of artefacts in the distorted image given by MAD. Brighter parts denote the higher visibility of the artefacts. \textbf{Lower row:} Gaussian smoothed image (of Kernel 10) of its upper corresponding image.}
\label{iqa4}
\end{figure*}

The other FR-IQA methods mainly rely on a global difference between the distorted image and its ground truth. 
Let us consider for example SSIM, MAD, FSIM, and GMSD. SSIM and MAD mainly compare the similarity of luminance and contrast between the distorted image and the ground truth. The other two, GSIM and GMSD, are based on the similarity of gradient magnitude. As shown in Fig.\ \ref{iqa4}, the main regions of the errors extracted by these four methods are to some extent similar and consistent with the most noticeable portion as shown in Fig.\ \ref{maindeg}. However, they still could not estimate the perceptual quality of motion-compensated interpolated images well (giving an average SROCC of 0.5370, 0.5220, 0.5339 and 0.5383, respectively). This may be caused by localized distortions, while FR-IQA methods typically make comparisons globally.

To overcome this problem, we propose a method based on weighted absolute error (WAE-IQA). It  computes the differences between an interpolated image and its ground truth for each pixel. Only pixels with a sufficiently large interpolation error are fully weighted and pixels with small errors are discounted. In addition a mild nonlinear scaling allows to shape weighted average error for better performance.

\begin{figure*}[h!]
\centering{\includegraphics[width=1\textwidth]{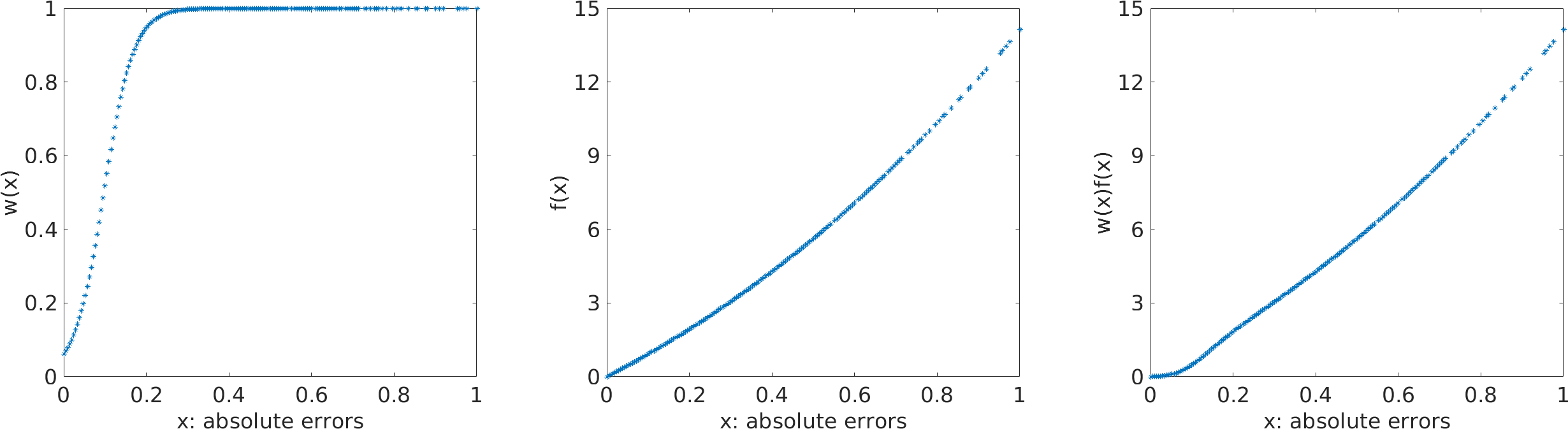}}
\vspace{-5pt}
\caption{Related functions in WAE-IQA of \emph{Dumptruck} with parameters $a_1= 8.7285, a_2=4.6443, a_3=0.7516, s= 28.0186, t=0.0973$. Left: logistic weight function $w(x)=1/( 1+  e^{-s(x -t)} )$. Middle: Polynomial function of absolute error $f(x)= a_1x + a_2x^2+a_3x^3$. Right: Weighted absolute error function $ w(x)f(x) $. }
\label{weight_abs}
\end{figure*}

\begin{figure*}[h!]
\centering{\includegraphics[width=1\textwidth]{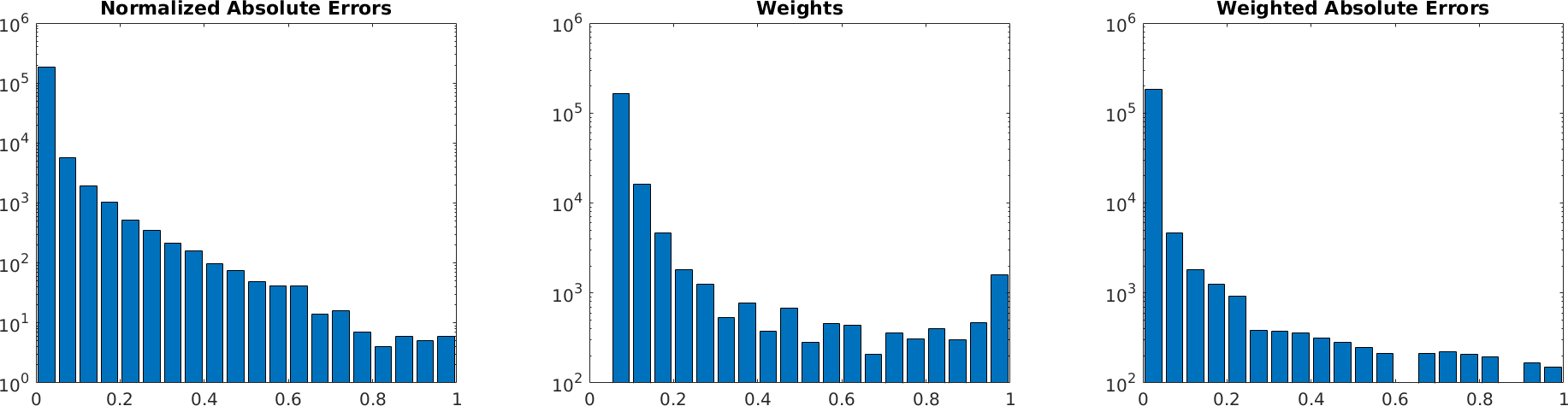}}
\vspace{-5pt}
\caption{Histogram of related values in WAE-IQA of \emph{Dumptruck} with parameters $a_1= 8.7285, a_2=4.6443, a_3=0.7516, s= 28.0186, t=0.0973$.} 
\label{wae_hist}
\end{figure*}

\begin{figure*}[h!]
\centering{\includegraphics[width=1\textwidth]{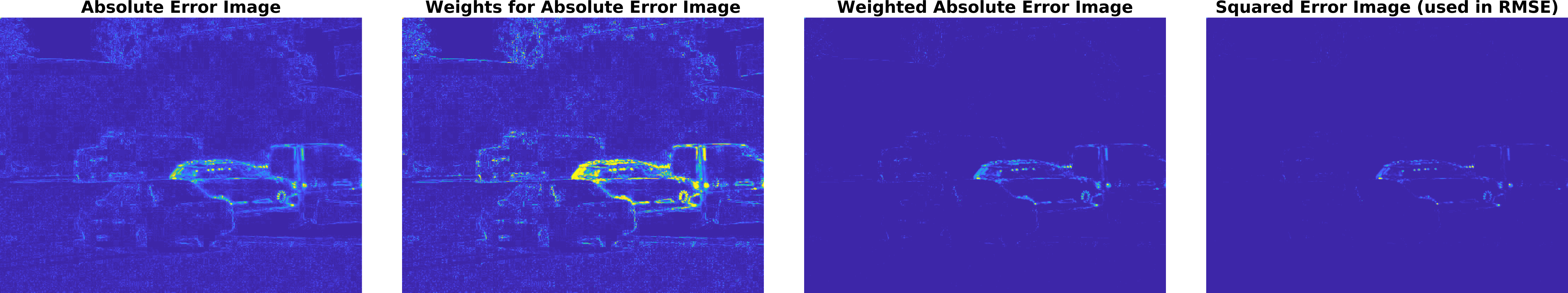}}
\vspace{-5pt}
\caption{Visualization of WAE for \emph{Dumptruck} with parameters  and squared errors used in RMSE.}
\label{vis_abs_dump}
\end{figure*}

The proposed method WAE proceeds as follows (see Fig.\ \ref{weight_abs}, \ref{wae_hist}, and \ref{vis_abs_dump}). 
\begin{itemize}
\item Images are converted from RGB to 8-bit grayscale, giving an interpolated grayscale image $\hat I$ and the corresponding grayscale ground truth image $I$.
\item The pixelwise absolute errors of $\hat I$ are computed and normalized to 1, which gives the normalized absolute error image $I'$: 
$$ I' = \frac{1}{255}|\hat{I}-I|.$$ 
    \item For normalized absolute errors $x \in [0,1]$ we define a weight $w(x) \in [0,1]$ by the logistic function 
    $$w(x)=\frac{1}{ 1+  e^{-s(x -t)}},$$
where the slope $s \in [0, \infty)$ and the shift $t \in [0,1]$ are parameters to be chosen.
    \item Normalized absolute errors $x \in [0,1]$ also nonlinearly scaled using a polynomial of degree 3, 
    $$f(x)= a_1x + a_2x^2+a_3x^3,$$ 
    where $a_1,a_2,a_3 \in [0,\infty)$ are parameters to be chosen.
    \item The weighted absolute error is
\begin{ceqn}
\begin{align}
WAE = \frac{\sum_{\text{pixel errors~} x} w(x)f(x)}{\sum_{\text{pixel errors~} x} w(x)}.
\end{align}
\end{ceqn}
\end{itemize}

 We selected the parameters $s,t,a_1, a_2, a_3$ optimally in terms of SROCC on the training sets in an 8-fold leave-one-out (LOO) cross-validation. We made use of the $ 155 \times 8$ interpolated images in the Middlebury benchmark together with their MOS obtained in our subjective study. For each LOO cross-validation, we used 7 sets for training, and the other set as the test set. 
 
 The SROCC results on the test set for each cross-validation is shown in Table \ref{tb:crd-iqa}. On average, WAE-IQA performed slightly better than RMSE. WAE-IQA performed best among all tested FR-IQA methods on three out of eight sets.

\section{Conclusion and Future Work}
We have adopted a well designed visual quality assessment to the Middlebury benchmark for frame interpolation mostly based on optical flow methods. Using artefact amplification, the sensitivity of our subjective study was increased significantly. Our study confirms that only using RMSE as an evaluation metric for image interpolation performance is not representative of subjective visual quality. Also current FR-IQA methods do not provide satisfying results on those interpolated images. This is due to the fact that such images, especially the ones generated by optical flow algorithms have specific distortions that are quite different from artefacts commonly addressed by conventional IQA methods. 
Hence, we proposed a novel FR-IQA method based on a weighted absolute error (WAE-IQA). While this approach outperformed the best FR-IQA method from the literature, the absolute ranking performance only improved slightly.
This illustrates that the quality assessment for motion-compen\-sated frame interpolation is a difficult task and there is still plenty of room for improvement.

In the future, we plan to make use of flow fields as side information to further improve the performance of evaluating the perceptual quality of motion-compensated interpolated images. Besides, in order to deal with the specific artefacts caused by motions in such images, we will investigate adopting video saliency detection methods as well as video quality assessment methods based on spatio-temporal information on such images. Furthermore, we will investigate adopting artefact amplification to I/VQA methods in order to improve their performances universally.

\section*{Conflict of interest}

The authors declare that they have no conflict of interest.


\bibliographystyle{spmpsci}
\bibliography{bib}

\end{document}